# Convergence and stability of Q-learning in Hierarchical Reinforcement Learning

Massimiliano Manenti*, Andrea Iannelli

*University of Stuttgart, Institute for Systems Theory and Automatic Control, Pfaffenwaldring 9, Stuttgart, 70569, Baden-Württemberg, Germany*

## HIGHLIGHTS

- Proposes Feudal Q-learning with coupled two-timescale updates.
- Provides the first proof of convergence and stability for interdependent HRL.
- Links HRL and Two-Timescale Stochastic Approximation theory.
- Interpretation of the algorithm's results as Nash and Stackelberg solutions.

## ABSTRACT

Hierarchical Reinforcement Learning promises, among other benefits, to efficiently capture and utilize the temporal structure of a decision-making problem and to enhance transfer learning capabilities, but theoretical guarantees lag behind practice. In this paper, we propose a Feudal Q-learning scheme and investigate under which conditions its coupled updates converge and are stable. By leveraging the theory of Stochastic Approximation and the ODE method, we present a theorem stating the convergence and stability properties of Feudal Q-learning. This provides a principled convergence and stability analysis tailored to Feudal RL. Moreover, we show that the updates converge to a point that can be interpreted as an equilibrium of a suitably defined game, opening the door to game-theoretic approaches to Hierarchical RL. Lastly, experiments based on the Feudal Q-learning algorithm support the outcomes anticipated by theory. Code and data available at: https://github.com/col-tasas/2025-convergence-and-stability-HRL.

## 1. Introduction

Decision-making architectures have played a central role for decades [1] both in engineering and in other domains, e.g., guidance, navigation and control of Apollo missions [2], chemical plants [3], smart grids [4], unmanned aerial vehicles [5], recommender systems [6], and algorithms [7]. Moreover, architectures are ubiquitous in nature; for example, diversity in the nervous system enables humans to have fast and accurate sensorimotor control [8]. Reinforcement Learning (RL) is a framework in which an agent learns to make sequential decisions through interaction with an environment in order to maximize cumulative reward [9]. Decision-making architectures have also been proposed and studied in RL. Hierarchical Reinforcement Learning (HRL) is a subfield of RL that deals with hierarchical structures for decision-making agents. Prospective advantages include improved long-term credit assignment, transfer learning, interpretability, and the integration of preexisting policies [10,11]. Many approaches have been proposed in HRL, e.g., Feudal RL [12,13], Options [14,15], MAXQ [16], and HAM [17]. Moreover, HRL has been shown to be effective in practice [13,18–21]. Despite the increasing interest in the theoretical understanding of HRL [22,23], the theoretical foundations of HRL remain underdeveloped compared with those of RL. Moreover, it is known that in HRL, non-stationarity issues can arise [18,19]. The main goal of our work is to contribute to the Feudal RL theory by proposing a Feudal Q-learning algorithm with proven theoretical guarantees.

In a two-level Feudal RL algorithm, two policies are interacting with each other. The high-level policy learns which goals to set for the low-level policy, which, in turn, learns how to achieve them. Broadly speaking, Feudal Q-learning consists of two interconnected Q-learning update rules that influence each other using data generated from a

* Corresponding author.
*Email addresses:* massimiliano.manenti@ist.uni-stuttgart.de (M. Manenti), andrea.iannelli@ist.uni-stuttgart.de (A. Iannelli).

Open access funding enabled and organized by Projekt DEAL.

stochastic process. To rigorously analyze such intertwined updates under stochastic data, we rely on the Stochastic Approximation (SA) framework [24,25], and in particular on its Two-Timescale variant (TTSA) [26]. By leveraging the ODE method [26,27], we study the update rules as coupled dynamical systems, which directly tackles the non-stationarity problem. We consider infinite-horizon Markov Decision Processes (MDPs) with finite state and action spaces and, by analyzing Feudal Q-learning through the lens of TTSA, we are able to prove the convergence and stability of the updates. To the best of the authors' knowledge, our work is the first to prove stability of an HRL algorithm and convergence for Feudal Q-learning updates that are interdependent. The closest works in the literature are [28,29]. In [28], the authors define a high-level infinite-horizon MDP and a low-level episodic MDP that is independent of the high-level policy. Assuming convergence in the low-level MDP, convergence in the high-level MDP is proved [[28], Theorem 2]. Our work differs from [28] in that two interconnected infinite-horizon MDPs are considered, and a coupled two-timescale analysis is performed, showing both convergence and stability of the updates. Our setup can bring some benefits, for example, it is easy to consider an extrinsic reward in the low-level infinite-horizon problem, which is shown to be beneficial in practice [13]. Moreover, in our framework, we do not need to assume that the duration of the low-level episode is sampled from a geometric distribution, which can potentially result in arbitrarily long episodes. Another drawback of the setup in [28] is that the choice of the low-level discount factor is tied to the choice of the expected episode duration. In [29], the work in [28] is extended by considering the convergence of Q-learning deep RL methods with non-stationary features that converge.

Another benefit of the approach we propose is that, by directly linking Feudal RL to the theory of TTSA, one could leverage results in the vast literature of TTSA, e.g., finite-time guarantees [30,31], or multi-timescales [32] which would enable deeper hierarchies. Thus, TTSA theory is not only employed to establish convergence and stability, but also to introduce a new perspective on HRL, by demonstrating that a Feudal RL problem can be rigorously cast within the TTSA framework. This is an important step toward closing the gap between theory and practice in HRL, as viewing the HRL problem as a TTSA problem can help in designing new and efficient algorithms with theoretical guarantees.

An additional contribution of our work is to formalize a game-theoretic interpretation of this HRL problem. Game theory and RL are closely related fields, as both deal with optimal decision-making. Probably the most popular connection is between game theory and multi-agent RL (MARL) [33], as MARL involves multiple agents interacting in a shared environment. It is interesting to note a more subtle connection between HRL and game theory: even though the original problem setup involves finding only one agent, we are decomposing the original MDP into multiple levels, and thus we can see each level as an agent in a multi-agent system. We highlight that our analysis shows the updates converge to a point that can be interpreted as both a Nash and a Stackelberg equilibrium, opening the door to game-theoretic approaches to HRL.

The contributions of this work can be summarized as follows:

- We propose a Feudal Q-learning algorithm consisting of two interacting infinite-horizon MDPs and cast its analysis within the TTSA framework.
- By establishing a rigorous connection between Feudal RL and TTSA, our framework enables the use of existing TTSA tools, including finite-time analyses and multi-timescale extensions for deeper hierarchies.
- We prove convergence and stability of the coupled Q-learning updates, directly addressing the non-stationarity arising in Hierarchical Reinforcement Learning.
- We provide a game-theoretic interpretation of the hierarchical interaction, showing that the limiting point can be viewed as a Nash and a Stackelberg equilibrium.

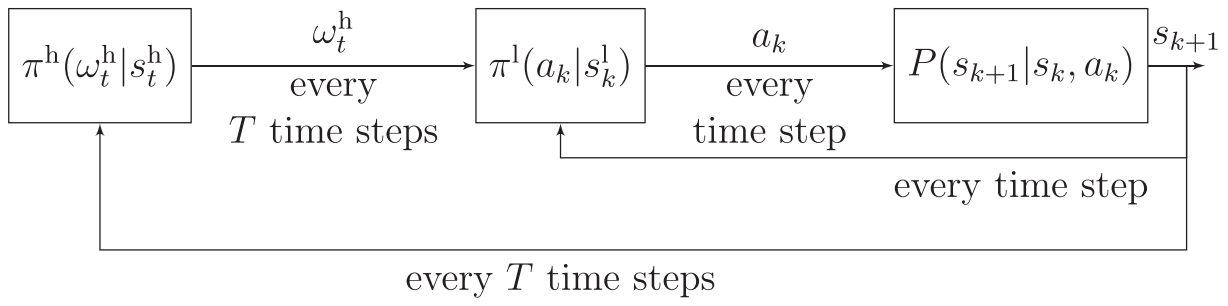

**Fig. 1.** Block diagram of the Feudal RL scheme. Instead of having a flat policy interacting with the system, there is a high-level policy, $\pi^{\mathrm{h}}$, selecting goals for a low-level policy, $\pi^{\mathrm{l}}$, that interacts with the system $P$.

- We perform numerical experiments to validate the theoretical results in practice.

*Notation.* Consider a function $f$ defined over a finite set $X$, we indicate with $\|\cdot\|_\infty$ the infinity norm, i.e., $\|f\|_\infty = \max_{x\in X}\{|f(x)|\}$ and with $\|\cdot\|_2$ the 2-norm, i.e., $\|f\|_2 = \sqrt{\sum_{x\in X}\left(f(x)\right)^2}$. The floor function $\lfloor\cdot\rfloor : \mathbb{R} \to \mathbb{Z}$ is defined as $\lfloor x \rfloor = \max\{n \in \mathbb{Z} | n \le x\}$. Consider an $m \times n$ matrix $A$ with the element $a_{ij}$ in the $i$-th row and $j$-th column, we indicate with $\mathrm{vec}(\cdot)$ the vectorization of the matrix obtained by stacking its columns, i.e., $[a_{1,1}, \dots, a_{m,1}, \dots, a_{1,n}, \dots, a_{m,n}]^\top = \mathrm{vec}(A)$.

## 2. Problem formulation

Consider an MDP, $\mathcal{M} = \langle \mathcal{S}, \mathcal{A}, P, r, \gamma \rangle$, where $\mathcal{S}$ is a finite state space, $\mathcal{A}$ is a finite action space, $P(s_{k+1}|s_k, a_k) : \mathcal{S}\times\mathcal{A}\times\mathcal{S} \to [0,1]$ is the dynamics, defining the probability of transitioning to the next state $s_{k+1} \in \mathcal{S}$ given the current state $s_k \in \mathcal{S}$ and action $a_k \in \mathcal{A}$, $r : \mathcal{S}\times\mathcal{A} \to \mathbb{R}$ is the deterministic reward function, and $\gamma \in (0,1)$ is the discount factor. A general goal in RL is to find a policy $\pi : \mathcal{S}\times\mathcal{A} \to [0,1]$ that maximizes, for every state $s$, $\mathbb{E}_{a\sim\pi}[Q_\pi(s,a)]$, where $Q_\pi : \mathcal{S}\times\mathcal{A} \to \mathbb{R}$ is defined as:

$$Q_\pi(s,a) := \lim_{N\to\infty} \mathbb{E}\left[\sum_{k=0}^{N} \gamma^k r(s_k, a_k) \middle| s_0 = s, a_0 = a\right], \tag{1}$$

and the expectation is taken according to $s_{k+1} \sim P(\cdot|s_k, a_k)$, $a_k \sim \pi(\cdot|s_k)$.

Feudal RL solves the MDP by decomposing the flat policy $\pi$ into a hierarchical structure, see Fig. 1. To mathematically formalize the problem, we propose two fictitious MDPs: the high-level MDP, $\mathcal{M}^{\mathrm{h}} := \langle \mathcal{S}^{\mathrm{h}}, \Omega, P^{\mathrm{h}}_{\pi^{\mathrm{l}}}, r^{\mathrm{h}}_{\pi^{\mathrm{l}}}, \gamma^{\mathrm{h}} \rangle$, and the low-level MDP, $\mathcal{M}^{\mathrm{l}} := \langle \mathcal{S}^{\mathrm{l}}, \mathcal{A}, P^{\mathrm{l}}_{\pi^{\mathrm{h}}}, r^{\mathrm{l}}, \gamma^{\mathrm{l}} \rangle$. Every $T$ time steps, the high-level policy, $\pi^{\mathrm{h}}$, selects a goal, $\omega_t^{\mathrm{h}} \in \Omega$, that the low-level policy, $\pi^{\mathrm{l}}$, should achieve. To achieve the goal, the low-level policy interacts with the system at every time step by selecting actions $a_k \in \mathcal{A}$.

A few general considerations on $\mathcal{M}^{\mathrm{h}}$ and $\mathcal{M}^{\mathrm{l}}$ are in order. Firstly, the designer has the freedom of choice in the goal space $\Omega$ and in the low-level reward $r^{\mathrm{l}}$. For example, $\Omega$ can coincide with $\mathcal{S}$ (or, more generally, a subset of the power set of $\mathcal{S}$) and $r^{\mathrm{l}}$ can be a combination of $r$ and a distance function describing how close the system is to the goal. While there is some freedom of choice in $r^{\mathrm{h}}_{\pi^{\mathrm{l}}}$ as well, it is often convenient to define it as a function of the extrinsic reward $r$. Secondly, elements of one MDP, e.g., $P^{\mathrm{h}}_{\pi^{\mathrm{l}}}$ and $r^{\mathrm{h}}_{\pi^{\mathrm{l}}}$, depend on the policy of the other MDP. This interdependence is key to capturing the non-stationarity issues arising in HRL. Lastly, $\mathcal{M}^{\mathrm{h}}$ and $\mathcal{M}^{\mathrm{l}}$ are running in parallel with potentially different time-scales: one time step in the high-level MDP corresponds to $T$ time steps in the low-level MDP. For this reason, we use $t$ to indicate the time step of the high-level MDP and $k$ for the time step of the low-level MDP, as shown in Fig. 2.

Two standard assumptions in theoretical RL, the boundedness of the rewards and the finiteness of the sets [34,35], are needed to prove the convergence of the proposed Feudal Q-learning updates.

**Assumption 1 (Bounded rewards).** The reward function $r$ and the low-level reward $r^{\mathrm{l}}$ are bounded, i.e., $\exists\, \bar{r}, \bar{r}^{\mathrm{l}} > 0$ such that $\|r(s,a)\|_\infty \le \bar{r}$ and $\left\|r^{\mathrm{l}}(s^{\mathrm{l}}, a)\right\|_\infty \le \bar{r}^{\mathrm{l}}$.

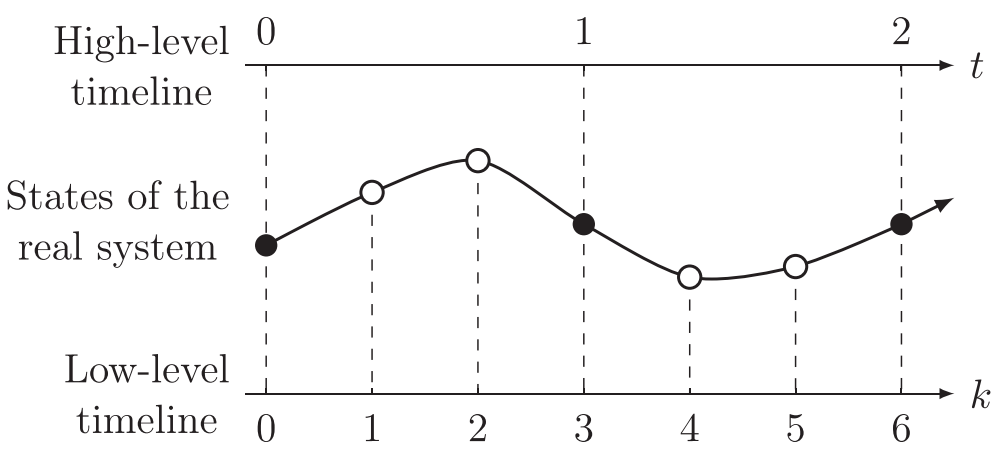

**Fig. 2.** Time scales of the high-level and low-level MDPs. Circles represent the states of the real system, while black circles indicate the states where the high-level MDP makes a decision. In this example, the high-level MDP makes a decision every $T = 3$ time steps of the low-level MDP.

**Assumption 2 (Finite sets).** The sets $\mathcal{S}$, $\mathcal{A}$, and $\Omega$ are finite.

### 2.1. High-level MDP

Let us now introduce the quantities comprising the high-level MDP $\mathcal{M}^{\mathrm{h}} = \langle \mathcal{S}^{\mathrm{h}}, \Omega, P^{\mathrm{h}}_{\pi^{\mathrm{l}}}, r^{\mathrm{h}}_{\pi^{\mathrm{l}}}, \gamma^{\mathrm{h}} \rangle$. We choose $\mathcal{S}^{\mathrm{h}} = \mathcal{S}$, and, for ease of exposition and without loss of generality, $\Omega = \mathcal{S}$. The variable $s^{\mathrm{h}}_t \in \mathcal{S}^{\mathrm{h}}$ represents the state of the high-level MDP at time $t$ and corresponds to the original MDP state $s_{tT}$, whereas $\omega^{\mathrm{h}}_t \in \Omega$ is the action of the high-level MDP at time $t$, i.e., the goal for the low-level MDP. $\pi^{\mathrm{h}} : \mathcal{S}^{\mathrm{h}} \times \Omega \to [0,1]$ represents the high-level policy that selects goals. $P^{\mathrm{h}}_{\pi^{\mathrm{l}}}(s^{\mathrm{h}}_{t+1} | s^{\mathrm{h}}_t, \omega^{\mathrm{h}}_t)$ and $r^{\mathrm{h}}_{\pi^{\mathrm{l}}}(s^{\mathrm{h}}_t, \omega^{\mathrm{h}}_t)$ describe, respectively, the dynamics and the reward in the high-level MDP. Before describing in more detail $P^{\mathrm{h}}_{\pi^{\mathrm{l}}}$ and $r^{\mathrm{h}}_{\pi^{\mathrm{l}}}$, let us introduce some useful quantities.

A decision epoch is a time step in the low-level MDP when the high-level MDP has made a decision, i.e., those low-level time steps belonging to the set $\{0, T, 2T, \ldots\}$. In order to take into account the time spent between decision epochs, we introduce the discrete variable $s^{\mathrm{de}}_k \in \mathcal{S}^{\mathrm{de}}$, where $\mathcal{S}^{\mathrm{de}} := \{0, \ldots, T-1\}$. The variable $s^{\mathrm{de}}_k$ is recursively defined as:

$$s^{\mathrm{de}}_{k+1} = \begin{cases} s^{\mathrm{de}}_k + 1, & \text{if } s^{\mathrm{de}}_k \neq T-1 \\ 0, & \text{if } s^{\mathrm{de}}_k = T-1 \end{cases}, \quad \text{with } s^{\mathrm{de}}_0 = 0.$$

Given the deterministic evolution of $s^{\mathrm{de}}$, one can easily find the probability $P^{\mathrm{de}}(s^{\mathrm{de}}_{k+1} | s^{\mathrm{de}}_k)$ of transitioning to $s^{\mathrm{de}}_{k+1}$ given $s^{\mathrm{de}}_k$. At time step $k$, the low-level policy $\pi^{\mathrm{l}} : \mathcal{S}^{\mathrm{l}} \times \mathcal{A} \to [0,1]$ selects an action $a_k \in \mathcal{A}$ to apply to the real system. We are now ready to present $P^{\mathrm{h}}_{\pi^{\mathrm{l}}}$ and $r^{\mathrm{h}}_{\pi^{\mathrm{l}}}$.

The high-level dynamics are:

$$P^{\mathrm{h}}_{\pi^{\mathrm{l}}}\left(s^{\mathrm{h}}_{t+1} | s^{\mathrm{h}}_t, \omega^{\mathrm{h}}_t\right) = \sum_{\substack{s_1,\ldots,s_{T-1} \in \mathcal{S} \\ a_0,\ldots,a_{T-1} \in \mathcal{A}}} \prod_{i=1}^{T} P(s_i | s_{i-1}, a_{i-1}) \pi^{\mathrm{l}}\left(a_{i-1} | s_{i-1}, \omega^{\mathrm{l}}_{i-1}, s^{\mathrm{de}}_{i-1}\right), \tag{2}$$

where $s_0 = s^{\mathrm{h}}_t$, $s_T = s^{\mathrm{h}}_{t+1}$, $\omega^{\mathrm{l}}_i = \omega^{\mathrm{h}}_t$, and $s^{\mathrm{de}}_i = i$, $\forall i \in \{0, \ldots, T-1\}$. The steps for obtaining Eq. (2) can be found in Appendix B. It is clear from Eq. (2) that the dynamics of the high-level MDP depend on the low-level policy, we highlight this dependence by placing the subscript $\pi^{\mathrm{l}}$ in $P^{\mathrm{h}}_{\pi^{\mathrm{l}}}$.

Since the high-level policy should learn how to solve the original MDP $\mathcal{M}$, the reward $r^{\mathrm{h}}_{\pi^{\mathrm{l}}}$ should depend on $r$, the reward of the original MDP. The high-level reward is:

$$r^{\mathrm{h}}_{\pi^{\mathrm{l}}}(s^{\mathrm{h}}, \omega^{\mathrm{h}}) = \mathop{\mathbb{E}}_{\pi^{\mathrm{l}}}\left[\sum_{k=0}^{T-1} (\gamma^{\mathrm{h}})^k r(s_k, a_k) | s_0 = s^{\mathrm{h}}\right], \tag{3}$$

where $\gamma^{\mathrm{h}} \in (0,1)$.

The high-level Q-function $Q^{\mathrm{h}}_{\pi^{\mathrm{h}},\pi^{\mathrm{l}}} : \mathcal{S}^{\mathrm{h}} \times \Omega \to \mathbb{R}$ is defined as:

$$Q^{\mathrm{h}}_{\pi^{\mathrm{h}},\pi^{\mathrm{l}}}(s^{\mathrm{h}}, \omega^{\mathrm{h}}) := \lim_{N \to \infty} \mathop{\mathbb{E}}_{\pi^{\mathrm{h}},\pi^{\mathrm{l}}}\left[\sum_{t=0}^{N} (\gamma^{\mathrm{h}})^{tT} r^{\mathrm{h}}_{\pi^{\mathrm{l}}}\left(s^{\mathrm{h}}_t, \omega^{\mathrm{h}}_t\right) \;\Big|\; s^{\mathrm{h}}_0 = s^{\mathrm{h}},\ \omega^{\mathrm{h}}_0 = \omega^{\mathrm{h}}\right]. \tag{4}$$

With the definitions of $r^{\mathrm{h}}_{\pi^{\mathrm{l}}}$ and $Q^{\mathrm{h}}_{\pi^{\mathrm{h}},\pi^{\mathrm{l}}}$, and by setting $\gamma^{\mathrm{h}} = \gamma$, if we consider a trajectory of the system, the experienced value of the high-level Q-function in Eq. (4) corresponds to that of the original MDP in Eq. (1). In this way, it is possible to compare the performance of a flat RL algorithm and a Feudal one.

### 2.2. Low-level MDP

Recall that the low-level MDP is $\mathcal{M}^{\mathrm{l}} := \langle \mathcal{S}^{\mathrm{l}}, \mathcal{A}, P^{\mathrm{l}}_{\pi^{\mathrm{h}}}, r^{\mathrm{l}}, \gamma^{\mathrm{l}} \rangle$. Here, the state space is $\mathcal{S}^{\mathrm{l}} = \mathcal{S} \times \Omega \times \mathcal{S}^{\mathrm{de}}$. The low-level state $s^{\mathrm{l}}_k := (s_k, \omega^{\mathrm{l}}_k, s^{\mathrm{de}}_k) \in \mathcal{S}^{\mathrm{l}}$ is composed of the state of the real system $s_k$, the goal to be tracked $\omega^{\mathrm{l}}_k = \omega^{\mathrm{h}}_{\lfloor k/T \rfloor}$, and $s^{\mathrm{de}}_k$. The low-level dynamics are:

$$P^{\mathrm{l}}_{\pi^{\mathrm{h}}}\left(s^{\mathrm{l}}_{k+1} | s^{\mathrm{l}}_k, a_k\right) = \\ = \begin{cases} 0, & \text{if } s^{\mathrm{de}}_k \neq T-1 \text{ and } \omega^{\mathrm{l}}_{k+1} \neq \omega^{\mathrm{l}}_k \\ P^{\mathrm{de}}\left(s^{\mathrm{de}}_{k+1} | s^{\mathrm{de}}_k\right) P(s_{k+1} | s_k, a_k), & \text{if } s^{\mathrm{de}}_k \neq T-1 \text{ and } \omega^{\mathrm{l}}_{k+1} = \omega^{\mathrm{l}}_k \\ P^{\mathrm{de}}\left(s^{\mathrm{de}}_{k+1} | s^{\mathrm{de}}_k\right) P(s_{k+1} | s_k, a_k) \pi^{\mathrm{h}}\left(\omega^{\mathrm{l}}_{k+1} | s_k\right), & \text{if } s^{\mathrm{de}}_k = T-1 \end{cases}. \tag{5}$$

The steps for obtaining Eq. (5) can be found in Appendix B. Analogously to the high-level dynamics, we also note that $P^{\mathrm{l}}_{\pi^{\mathrm{h}}}$ depends on the high-level policy.

As mentioned earlier, there is no fixed choice of the low-level reward $r^{\mathrm{l}}$, which is a free design parameter. The low-level Q-function $Q^{\mathrm{l}}_{\pi^{\mathrm{h}},\pi^{\mathrm{l}}} : \mathcal{S}^{\mathrm{l}} \times \mathcal{A} \to \mathbb{R}$ is defined as:

$$Q^{\mathrm{l}}_{\pi^{\mathrm{h}},\pi^{\mathrm{l}}}(s^{\mathrm{l}}, a) := \lim_{N \to \infty} \mathop{\mathbb{E}}_{\pi^{\mathrm{h}},\pi^{\mathrm{l}}}\left[\sum_{k=0}^{N} (\gamma^{\mathrm{l}})^k r^{\mathrm{l}}\left(s^{\mathrm{l}}_k, a_k\right) \;\Big|\; s^{\mathrm{l}}_0 = s^{\mathrm{l}},\ a_0 = a\right], \tag{6}$$

where $\gamma^{\mathrm{l}} \in (0,1)$.

### 2.3. Problem statement

Having defined the high- and low-level MDPs, we can now state the problem we want to solve. Our goal is to find the optimal high- and low-level Q-functions, $Q^{\mathrm{h},*} : \mathcal{S}^{\mathrm{h}} \times \Omega \to \mathbb{R}$ and $Q^{\mathrm{l},*} : \mathcal{S}^{\mathrm{l}} \times \mathcal{A} \to \mathbb{R}$, that satisfy, for every state-action pair, the Bellman equations:

$$Q^{\mathrm{h},*}\left(s^{\mathrm{h}}_t, \omega^{\mathrm{h}}_t\right) = r^{\mathrm{h}}_{\pi^{\mathrm{l},*}}\left(s^{\mathrm{h}}_t, \omega^{\mathrm{h}}_t\right) + (\gamma^{\mathrm{h}})^T \mathop{\mathbb{E}}_{\pi^{\mathrm{l},*}}\left[\max_{\omega^{\mathrm{h}}_{t+1}} Q^{\mathrm{h},*}\left(s^{\mathrm{h}}_{t+1}, \omega^{\mathrm{h}}_{t+1}\right)\right], \tag{7a}$$

$$Q^{\mathrm{l},*}\left(s^{\mathrm{l}}_k, a_k\right) = r^{\mathrm{l}}\left(s^{\mathrm{l}}_k, a_k\right) + \gamma^{\mathrm{l}} \mathop{\mathbb{E}}_{\pi^{\mathrm{h},*}}\left[\max_{a_{k+1}} Q^{\mathrm{l},*}\left(s^{\mathrm{l}}_{k+1}, a_{k+1}\right)\right], \tag{7b}$$

$$\text{with } \pi^{\mathrm{h},*} \in \arg\max_{\pi^{\mathrm{h}}} Q^{\mathrm{h}}_{\pi^{\mathrm{h}},\pi^{\mathrm{l},*}} \text{ and } \pi^{\mathrm{l},*} \in \arg\max_{\pi^{\mathrm{l}}} Q^{\mathrm{l}}_{\pi^{\mathrm{h},*},\pi^{\mathrm{l}}}. \tag{7c}$$

It is interesting to note that the Bellman equation for one level, e.g., Eq. (7a), depends on the policy of the other level, e.g., $\pi^{\mathrm{l},*}$. This is a non-stationarity issue typical of HRL algorithms that complicates their analysis [18,19].

The goal that we want to reach is to jointly learn optimal Q-functions, while the other level uses the optimal policy.

Table 1 summarizes the main variables and symbols presented so far.

## 3. Convergence and stability guarantees of Feudal Q-learning

In this section, in order to solve the problem in Eq. (7), we will propose an update rule based on Q-learning. Since the TTSA theory constitutes our primary theoretical framework, we will briefly recall its main elements and employ it to derive insights into the behaviour of the proposed Feudal Q-learning updates. The Feudal Q-learning updates, $Q^{\mathrm{h},n}$ and $Q^{\mathrm{l},n}$, where $n \geq 0$ represents the iteration, will be shown to converge to the optimal high- and low-level Q-functions, $Q^{\mathrm{h},*}$ and $Q^{\mathrm{l},*}$ as defined in Eq. (7). Moreover, we will prove that the learning process is stable, and thus $Q^{\mathrm{h},n}$ and $Q^{\mathrm{l},n}$ will always have finite values.

**Table 1**
Table summarizing the main symbols.

| Symbol | Meaning |
|---|---|
| $a_k$ | Low-level action at time step $k$. |
| $\mathcal{A}$ | Action space of the original MDP. |
| $\gamma$, $\gamma^{\mathrm{h}}$, $\gamma^{\mathrm{l}}$ | Discount factors. |
| $\mathcal{M}$, $\mathcal{M}^{\mathrm{h}}$, $\mathcal{M}^{\mathrm{l}}$ | Original, high-, and low-level MDP. |
| $\omega_t^{\mathrm{h}}$ | High-level action at time step $t$. |
| $\omega_k^{\mathrm{l}}$ | Goal, at time step $k$, that the low-level policy should achieve. |
| $\Omega$ | Action space for the high-level MDP. |
| $P$, $P_{\pi^{\mathrm{l}}}^{\mathrm{h}}$, $P_{\pi^{\mathrm{h}}}^{\mathrm{l}}$ | Dynamics of the original, high-, and low-level MDP. |
| $P^{\mathrm{de}}$ | Dynamics of $s^{\mathrm{de}}$. |
| $\pi$, $\pi^{\mathrm{h}}$, $\pi^{\mathrm{l}}$ | Flat, high-, and low-level policy. |
| $\pi^{\mathrm{h},*}$, $\pi^{\mathrm{l},*}$ | Optimal high- and low-level policy (Eq. 7c). |
| $Q_\pi$ | Q-function of a (flat) policy $\pi$. |
| $Q_{\pi^{\mathrm{h}},\pi^{\mathrm{l}}}^{\mathrm{h}}$, $Q_{\pi^{\mathrm{h}},\pi^{\mathrm{l}}}^{\mathrm{l}}$ | High- and low-level Q-function. |
| $Q^{\mathrm{h},*}$, $Q^{\mathrm{l},*}$ | Optimal high- and low-level Q-function (Eq. 7). |
| $r$ | Reward of the original MDP. |
| $r_{\pi^{\mathrm{l}}}^{\mathrm{h}}$, $r^{\mathrm{l}}$ | High- and low-level reward. |
| $\bar{r}$, $\bar{r}^{\mathrm{l}}$ | Bounds on the reward functions. |
| $\mathcal{S}$, $\mathcal{S}^{\mathrm{h}}$, $\mathcal{S}^{\mathrm{l}}$ | Original, high-, and low-level state space. |
| $s_k$ | State of the original MDP at time step $k$. |
| $s_t^{\mathrm{h}}$ | High-level state at time $t$ (corresponds to $s_{tT}$). |
| $s_k^{\mathrm{l}}$ | Low-level state at time $k$ (corresponds to $(s_k, \omega_k^{\mathrm{l}}, s_k^{\mathrm{de}})$). |
| $s_k^{\mathrm{de}}$ | Variable considering the time spent between decision epochs. |
| $\mathcal{S}^{\mathrm{de}}$ | Set $\{0, \dots, T-1\}$ to which $s_k^{\mathrm{de}}$ belongs. |
| $T$ | Number of low-level time steps between consecutive high-level decisions. |

### *3.1. Feudal Q-learning updates*

As the Feudal Q-learning updates depend on trajectories collected through experiments or simulations, before presenting the update rule, we will make an assumption about how the data is collected.

**Assumption 3 (On-policy updates and Lipschitz continuous policies).** The policies, $\pi^{\mathrm{h},n}$ and $\pi^{\mathrm{l},n}$, that collect the trajectories used to update $Q^{\mathrm{h},n}$ and $Q^{\mathrm{l},n}$ depend respectively on $Q^{\mathrm{h},n}$ and $Q^{\mathrm{l},n}$, i.e., $\pi^{\mathrm{h},n} = \mathcal{F}^{\mathrm{h},n}(Q^{\mathrm{h},n})$ and $\pi^{\mathrm{l},n} = \mathcal{F}^{\mathrm{l},n}(Q^{\mathrm{l},n})$, where the operators $\mathcal{F}^{\mathrm{h},n} : \mathbb{R}^{|\mathcal{S}^{\mathrm{h}}|\times|\Omega|} \to \mathbb{R}^{|\mathcal{S}^{\mathrm{h}}|\times|\Omega|}$ and $\mathcal{F}^{\mathrm{l},n} : \mathbb{R}^{|\mathcal{S}^{\mathrm{l}}|\times|\mathcal{A}|} \to \mathbb{R}^{|\mathcal{S}^{\mathrm{l}}|\times|\mathcal{A}|}$ denote the policy extraction methods.

Moreover, the high-level policy is Lipschitz continuous with respect to $Q^{\mathrm{h},n}$, i.e., $\forall n$, $\exists L_{\pi^{\mathrm{h}}}^{Q^{\mathrm{h}}} \in \mathbb{R}_{\geq 0} : \forall (Q_1^{\mathrm{h}}, Q_2^{\mathrm{h}}) \in \mathbb{R}^{|\mathcal{S}^{\mathrm{h}}|\times|\Omega|} \times \mathbb{R}^{|\mathcal{S}^{\mathrm{h}}|\times|\Omega|}$,

$$|\mathcal{F}^{\mathrm{h},n}\left(Q_1^{\mathrm{h}}\right) - \mathcal{F}^{\mathrm{h},n}\left(Q_2^{\mathrm{h}}\right)| \leq L_{\pi^{\mathrm{h}}}^{Q^{\mathrm{h}}} \left\| Q_1^{\mathrm{h}} - Q_2^{\mathrm{h}} \right\|_\infty .$$

An analogous assumption holds for the low-level policy, with a constant $L_{\pi^{\mathrm{l}}}^{Q^{\mathrm{l}}}$.

Assumption 3 is not overly restrictive, for example, a Boltzmann exploration policy satisfies Assumption 3 [36]. However, it is worth noting that the standard $\epsilon$-greedy policy does not satisfy the Lipschitz continuity part of Assumption 3. For this reason, the numerical experiments in Section 5 will be performed with a Boltzmann exploration policy.

We are now ready to present the Feudal Q-learning updates:

$$Q^{\mathrm{h},n+1}\left(s_t^{\mathrm{h}}, \omega_t^{\mathrm{h}}\right) = Q^{\mathrm{h},n}\left(s_t^{\mathrm{h}}, \omega_t^{\mathrm{h}}\right) + \beta(n)\Bigg[\sum_{k=0}^{T-1}\left(\gamma^{\mathrm{h}}\right)^k r(s_{tT+k}, a_{tT+k}) + \left(\gamma^{\mathrm{h}}\right)^T \max_{\omega} Q^{\mathrm{h},n}\left(s_{t+1}^{\mathrm{h}}, \omega\right) - Q^{\mathrm{h},n}\left(s_t^{\mathrm{h}}, \omega_t^{\mathrm{h}}\right)\Bigg], \tag{8a}$$

$$Q^{\mathrm{l},n+1}\left(s_k^{\mathrm{l}}, a_k\right) = Q^{\mathrm{l},n}\left(s_k^{\mathrm{l}}, a_k\right) + \alpha(n)\left[r^{\mathrm{l}}\left(s_k^{\mathrm{l}}, a_k\right) + \gamma^{\mathrm{l}} \max_a Q^{\mathrm{l},n}\left(s_{k+1}^{\mathrm{l}}, a\right) - Q^{\mathrm{l},n}\left(s_k^{\mathrm{l}}, a_k\right)\right], \tag{8b}$$

with step-sizes $\alpha(n)$, $\beta(n) > 0$.

There is no requirement for how to initialize the iterative updates, e.g., $Q^{\mathrm{h},0}$ and $Q^{\mathrm{l},0}$ can be set to zero or they can be optimistically initialized [37].

### *3.2. TTSA and its connection to Feudal Q-learning*

Pioneered by [24], SA algorithms are a family of iterative methods to solve root finding problems, e.g., finding $z^* \in \mathbb{R}^d : f(z^*) = 0$, where only noisy data is available, e.g., we only have access to $f(z_n) + M_{n+1}$, where $M_{n+1}$ is a martingale difference sequence playing the role of the noise. A classic Single Timescale SA scheme can be written as:

$$z_{n+1} = z_n + \alpha(n)[f(z_n) + M_{n+1}], \tag{9}$$

where $z_n \in \mathbb{R}^d$ are the iterates of the algorithm. When referring to Two-Timescale SA schemes, the updates are:

$$x_{n+1} = x_n + \alpha(n)\left[h(x_n, y_n) + M_{n+1}^{(1)}\right], \tag{10a}$$

$$y_{n+1} = y_n + \beta(n)\left[g(x_n, y_n) + M_{n+1}^{(2)}\right], \tag{10b}$$

where $x_n$ and $y_n$ can be thought of as the Two-Timescale counterpart of $z_n$.

Two fundamental properties of an SA scheme are convergence and stability of the algorithm. Convergence refers to the iterates converging almost surely (a.s.) to $z^*$, while stability indicates that the iterates remain bounded (preferably without the need to project the iterates into a predefined set) [25]. A notable method to analyze the convergence and stability of SA schemes is the so-called ODE method [27]. The central idea is to show that the iterate $z_n$ "tracks" the continuous time ODE $\dot{z}(t) = f(z(t))$, whose solutions converge to $z^*$. The ODE method is supported by a rich theoretical foundation both for the Single Timescale case and the Two-Timescale case [25–27,38,39].

Concerning our Feudal Q-learning problem, we firstly cast the updates in Eq. (8) as Eq. (10), with $x_n$ and $y_n$ respectively representing $Q^{\mathrm{l},n}$ and $Q^{\mathrm{h},n}$. Secondly, similar to [25, Chapter 8], we study the related singularly perturbed ODEs:

$$\dot{x}(t) = \frac{1}{\epsilon} h(x(t), y(t)), \tag{11a}$$

$$\dot{y}(t) = g(x(t), y(t)), \tag{11b}$$

that the iterates $x_n$ and $y_n$ should track. The vector-valued function $h$ has as component $h_{s^{\mathrm{l}},a}(x_n, y_n)$, related to the low-level state and action pair $(s^{\mathrm{l}}, a)$, the low-level Bellman error:

$$h_{s^{\mathrm{l}},a}(x_n, y_n) := r^{\mathrm{l}}(s^{\mathrm{l}}, a) + \gamma^{\mathrm{l}} \mathop{\mathbb{E}}_{P_{\pi^{\mathrm{h}}}^{\mathrm{l}}}\left[\max_{a^+} Q^{\mathrm{l},n}(s^{\mathrm{l},+}, a^+)\right] - Q^{\mathrm{l},n}(s^{\mathrm{l}}, a), \tag{12}$$

while the vector-valued function $g$ has as component $g_{s^{\mathrm{h}},\omega^{\mathrm{h}}}(x_n, y_n)$, related to the high-level state and action pair $(s^{\mathrm{h}}, \omega^{\mathrm{h}})$, the high-level Bellman error:

$$g_{s^{\mathrm{h}},\omega^{\mathrm{h}}}(x_n, y_n) := r_{\pi^{\mathrm{l}}}^{\mathrm{h}}(s^{\mathrm{h}}, \omega^{\mathrm{h}}) + \left(\gamma^{\mathrm{h}}\right)^T \mathop{\mathbb{E}}_{P_{\pi^{\mathrm{l}}}^{\mathrm{h}}}\left[\max_{\omega^+} Q^{\mathrm{h},n}(s^{\mathrm{h},+}, \omega^+)\right] - Q^{\mathrm{h},n}(s^{\mathrm{h}}, \omega^{\mathrm{h}}). \tag{13}$$

For more details, see Appendix A. The parameter $\epsilon > 0$ is the singular perturbation parameter and, by having $\epsilon \downarrow 0$, we impose faster, $x(t)$, and slower, $y(t)$, dynamics. This timescale separation is obtained in the Feudal Q-learning algorithm by imposing the condition $\frac{\beta(n)}{\alpha(n)} \xrightarrow{n\to\infty} 0$ on the step-sizes, which motivates part of the following assumption; the remaining parts are standard Robbins-Monro conditions [24].

**Assumption 4 (Step-sizes).** The step-sizes satisfy the following conditions:

a) the step-sizes are not summable:

$$\lim_{N\to\infty}\sum_{n=0}^{N}\alpha(n)=\lim_{N\to\infty}\sum_{n=0}^{N}\beta(n)=\infty;$$

b) the step-sizes are square-summable:

$$\lim_{N\to\infty}\sum_{n=0}^{N}\left(\alpha(n)^2+\beta(n)^2\right)<\infty;$$

c) in the limit, the ratio of the step-sizes satisfies:

$$\lim_{n\to\infty}\frac{\beta(n)}{\alpha(n)}=0.$$

Within the TTSA framework, the timescale separation ensures that the iterates of the algorithm in Eq. (10) "track" the solutions of the following ODEs:

$$\frac{\mathrm{d}x}{\mathrm{d}t}=h(x(t),y), \tag{14}$$

where $y$ is held fixed as a constant parameter, and:

$$\frac{\mathrm{d}y}{\mathrm{d}t}=g(\lambda(y(t)),y(t)), \tag{15}$$

where $\lambda:\mathbb{R}^{|S^{\mathrm{h}}|\times|\Omega|}\to\mathbb{R}^{|S^{\mathrm{l}}|\times|\mathcal{A}|}$.

One can easily see from Eqs. (12) and (13) that equilibria of the ODEs (14) and (15) correspond to solutions of the Bellman equations. Thus, studying the asymptotic behaviour of the ODEs, as done in Lemmas 1 and 2, is of great interest and sheds light on the behaviour of the algorithm.

**Lemma 1 (Existence and properties of the globally asymptotically stable equilibrium of $h$).** *The ODE (14) has a globally asymptotically stable equilibrium $\lambda(y)$. Moreover, $\lambda$ is a Lipschitz map.*

**Lemma 2 (Existence of the globally asymptotically stable equilibrium of $g$).** *The ODE (15) has a globally asymptotically stable equilibrium $y^*$.*

For a proof of Lemmas 1 and 2, which will be used to prove Theorem 1, see Appendix C.

An interpretation of Lemma 1 is that, for a fixed high-level Q-function, $y$, the low-level Q-function, $x$, converges to the solution of the low-level Bellman equation. Thus, while the high-level Q-function is slowly varying, the low-level Q-function quickly converges to the optimal low-level Q-function. Even though it is quasi-static from the perspective of the low-level Q-function, the high-level Q-function is not constant. Lemma 2 tells us that the high-level Q-function converges to the solution of the high-level Bellman equation, where the low-level Q-function is assumed to always be at its optimal value.

### 3.3. Theoretical guarantees

After establishing the remaining assumptions, this subsection presents the main theoretical result of the paper.

Since we are using an on-policy algorithm, we would like to be greedy in the limit, in order to converge to the optimal policies. Furthermore, we would like to ensure infinite state-action visitation, which is a standard requirement in Q-learning [40]. For this reason, we require the following assumptions, which are stated for the sake of readability only for the high-level policy, but apply to the low-level one as well.

**Assumption 5 (Greedy in the limit policies).** Assume, for the sake of simplicity, that for a given state $s^{\mathrm{h}}$, $Q^{\mathrm{h},n}(s^{\mathrm{h}},\omega)$ has a unique maximizer $\omega^{\mathrm{h}}$. The high-level policy is greedy in the limit, i.e., as $n\to\infty$:

$$\pi^{\mathrm{h},n}(\omega^{\mathrm{h}}|s^{\mathrm{h}})=\begin{cases}1, & \omega^{\mathrm{h}}=\arg\max_{\omega\in\Omega}Q^{\mathrm{h},n}(s^{\mathrm{h}},\omega)\\ 0, & \text{otherwise}\end{cases}. \tag{16}$$

**Assumption 6 (Infinite exploration policies).** Consider a high-level trajectory $(s_t^{\mathrm{h}},\omega_t^{\mathrm{h}})_{t\geq0}$ and denote by $n_t(s^{\mathrm{h}},\omega^{\mathrm{h}})$ the number of times action $\omega^{\mathrm{h}}$ has been chosen in state $s^{\mathrm{h}}$ during the first $t$ time steps, i.e., $n_t(s^{\mathrm{h}},\omega^{\mathrm{h}}):=\sum_{i=0}^{t-1}\mathbb{1}(s_i^{\mathrm{h}}=s^{\mathrm{h}},\omega_i^{\mathrm{h}}=\omega^{\mathrm{h}})$, where $\mathbb{1}(\cdot)$ represents the indicator function. We assume that, $\forall(s^{\mathrm{h}},\omega^{\mathrm{h}})\in\mathcal{S}^{\mathrm{h}}\times\Omega$, $\lim_{t\to\infty}n_t(s^{\mathrm{h}},\omega^{\mathrm{h}})=\infty$ a.s.

Policies satisfying Assumptions 5 and 6 are referred to as Greedy in the Limit with Infinite Exploration (GLIE), see [41] for a Boltzmann exploration policy and an $\epsilon$-greedy policy that are GLIE.

An additional assumption is needed to prove the stability of the algorithm [38].

**Assumption 7 (Compact convergence).** The policies $\pi_c^{\mathrm{h}}:=\mathcal{F}^{\mathrm{h},n}(cQ^{\mathrm{h}})$, $c\geq1$, satisfy $\pi_c^{\mathrm{h}}\to\pi_\infty^{\mathrm{h}}$ as $c\to\infty$, uniformly on compacts for some $\pi_\infty^{\mathrm{h}}$.

See [38, Proposition 11] for a Boltzmann policy that satisfies Assumption 7. Moreover, an analogous assumption applies to the low-level policy.

Having laid out the necessary assumptions, we are now in a position to state the main theorem.

**Theorem 1 (Convergence and stability of Feudal Q-learning).** *If Assumptions 1–7 hold, then:*

*a) Feudal Q-learning converges to the optimal values:*

$$\lim_{n\to\infty}(Q^{h,n},Q^{l,n})=(Q^{h,*},Q^{l,*}),\ a.s.; \tag{17}$$

*b) The iterates $(Q^{h,n},Q^{l,n})$ are stable:*

$$\sup_n(||Q^{h,n}||+||Q^{l,n}||)<\infty,\ a.s. \tag{18}$$

**Proof.** See Appendix C. □

This result provides the first convergence and stability guarantees on Q-learning for Feudal RL. While relying on classical techniques used for analysis of RL algorithms [27,35,42], this result represents an important addition to the existing literature. Firstly, in order to leverage the existing theory on TTSA, the updates had to be cast as on-policy updates (Assumption 3). Otherwise, the terms $M_{n+1}^{(1)}$ and $M_{n+1}^{(2)}$ would not be martingale difference sequences, resulting in a non-vanishing bias. Secondly, having cast the Feudal Q-learning updates within the TTSA framework and interpreted the algorithm as coupled slow and fast updates, one can view the design of HRL algorithms as a problem of synthesizing the underlying ODEs. In particular, modifying the mean fields $h$ and $g$ or introducing structured gains [43] offers a principled route to derive variants with improved transient behaviour or asymptotic properties. Lastly, as reported in Section 3.2, our result sheds light on the behaviour of HRL algorithms. Qualitatively, faster updates in the low-level, which translate in our setup into bigger step-sizes for the low-level, are beneficial for convergence.

## 4. Game-theoretic interpretation

In this section, we highlight the connection between HRL and game theory. In particular, we focus on the proposed Feudal Q-learning scheme and discuss how it can be interpreted as a two-player game, whose solution is, in this special case, both a Nash equilibrium and a Stackelberg equilibrium.

Learning in the high- and low-level MDPs can be seen as a two-player game, where the high-level player is trying to maximize $Q^{\mathrm{h}}_{\pi^{\mathrm{h}},\pi^{\mathrm{l}}}$ and the low-level player is trying to maximize $Q^{\mathrm{l}}_{\pi^{\mathrm{h}},\pi^{\mathrm{l}}}$. The agents are learning Q-functions $Q^{\mathrm{h},n}$ and $Q^{\mathrm{l},n}$ and, through mappings $\mathcal{F}^{\mathrm{h},n}$ and $\mathcal{F}^{\mathrm{l},n}$, use them to derive their policies, which will be greedy in the limit as in Assumption 5. For this reason, with a slight abuse of notation, we sometimes replace the policy symbol with a Q-function.

We then define the optimal Bellman operators $\mathcal{T}^{\mathrm{h}}_{Q^{\mathrm{l}}} : \mathbb{R}^{|\mathcal{S}^{\mathrm{h}}|\times|\Omega|} \to \mathbb{R}^{|\mathcal{S}^{\mathrm{h}}|\times|\Omega|}$ and $\mathcal{T}^{\mathrm{l}}_{Q^{\mathrm{h}}} : \mathbb{R}^{|\mathcal{S}^{\mathrm{l}}|\times|\mathcal{A}|} \to \mathbb{R}^{|\mathcal{S}^{\mathrm{l}}|\times|\mathcal{A}|}$ as:

$$\left(\mathcal{T}^{\mathrm{h}}_{Q^{\mathrm{l}}} Q^{\mathrm{h}}\right)\left(s^{\mathrm{h}}_t, \omega^{\mathrm{h}}_t\right) := r^{\mathrm{h}}_{Q^{\mathrm{l}}}\left(s^{\mathrm{h}}_t, \omega^{\mathrm{h}}_t\right) + \left(\gamma^{\mathrm{h}}\right)^T \mathop{\mathbb{E}}_{Q^{\mathrm{l}}}\left[\max_{\omega^{\mathrm{h}}_{t+1}} Q^{\mathrm{h}}\left(s^{\mathrm{h}}_{t+1}, \omega^{\mathrm{h}}_{t+1}\right)\right], \quad \forall \left(s^{\mathrm{h}}_t, \omega^{\mathrm{h}}_t\right) \in \mathcal{S}^{\mathrm{h}} \times \Omega, \tag{19}$$

$$\left(\mathcal{T}^{\mathrm{l}}_{Q^{\mathrm{h}}} Q^{\mathrm{l}}\right)\left(s^{\mathrm{l}}_k, a_k\right) := r^{\mathrm{l}}\left(s^{\mathrm{l}}_k, a_k\right) + \gamma^{\mathrm{l}} \mathop{\mathbb{E}}_{Q^{\mathrm{h}}}\left[\max_{a_{k+1}} Q^{\mathrm{l}}\left(s^{\mathrm{l}}_{k+1}, a_{k+1}\right)\right], \quad \forall \left(s^{\mathrm{l}}_k, a_k\right) \in \mathcal{S}^{\mathrm{l}} \times \mathcal{A}. \tag{20}$$

A standard result in RL states that a greedy policy with respect to the optimal Q-function, i.e., the Q-function that is a fixed point of the optimal Bellman operator, represents an optimal policy [9, Chapter 3]. This motivates the following definitions [44]. We call high-level rational reaction set, $D^{\mathrm{h}}$, and low-level rational reaction set, $D^{\mathrm{l}}$, the sets:

$$D^{\mathrm{h}} := \left\{ (Q^{\mathrm{h}}, Q^{\mathrm{l}}) \in \mathbb{R}^{|\mathcal{S}^{\mathrm{h}}|\times|\Omega|} \times \mathbb{R}^{|\mathcal{S}^{\mathrm{l}}|\times|\mathcal{A}|} \mid Q^{\mathrm{h}} = \mathcal{T}^{\mathrm{h}}_{Q^{\mathrm{l}}} Q^{\mathrm{h}} \right\}, \tag{21}$$

$$D^{\mathrm{l}} := \left\{ (Q^{\mathrm{h}}, Q^{\mathrm{l}}) \in \mathbb{R}^{|\mathcal{S}^{\mathrm{h}}|\times|\Omega|} \times \mathbb{R}^{|\mathcal{S}^{\mathrm{l}}|\times|\mathcal{A}|} \mid Q^{\mathrm{l}} = \mathcal{T}^{\mathrm{l}}_{Q^{\mathrm{h}}} Q^{\mathrm{l}} \right\}. \tag{22}$$

For completeness, we introduce the definitions of Nash equilibrium and Stackelberg equilibrium [44,45].

**Definition 1 (Nash equilibrium).** A pair $(\bar{Q}^{\mathrm{h}}, \bar{Q}^{\mathrm{l}})$ such that $(\bar{Q}^{\mathrm{h}}, \bar{Q}^{\mathrm{l}}) \in D^{\mathrm{h}} \cap D^{\mathrm{l}}$ is a Nash equilibrium.

**Definition 2 (Stackelberg equilibrium).** A pair $(\bar{Q}^{\mathrm{h}}, \bar{Q}^{\mathrm{l}}) \in D^{\mathrm{l}}$ such that

$$Q^{\mathrm{h}}_{\bar{Q}^{\mathrm{h}},\bar{Q}^{\mathrm{l}}} \geq Q^{\mathrm{h}}_{Q^{\mathrm{h}},Q^{\mathrm{l}}}, \quad \forall (Q^{\mathrm{h}}, Q^{\mathrm{l}}) \in D^{\mathrm{l}}, \tag{23}$$

is a Stackelberg equilibrium with player $\bar{Q}^{\mathrm{h}}$ as a leader and player $\bar{Q}^{\mathrm{l}}$ as a follower.

Now, we are ready to state and prove Lemma 3, connecting our HRL updates to game theory.

**Lemma 3.** *The pair $(Q^{\mathrm{h},*}, Q^{\mathrm{l},*})$ found by the Feudal Q-learning updates is both a Nash equilibrium and a Stackelberg equilibrium where the high-level is the leader and the low-level is the follower.*

**Proof.** In order to show that $(Q^{\mathrm{h},*}, Q^{\mathrm{l},*})$ is a Nash equilibrium, we need to show that $(Q^{\mathrm{h},*}, Q^{\mathrm{l},*}) \in D^{\mathrm{h}} \cap D^{\mathrm{l}}$. It is immediate to verify this by noting that $Q^{\mathrm{h},*}$ and $Q^{\mathrm{l},*}$ satisfy Eqs. (7a) and (7b), i.e., they verify the conditions $Q^{\mathrm{h},*} = \mathcal{T}^{\mathrm{h}}_{Q^{\mathrm{l},*}} Q^{\mathrm{h},*}$ and $Q^{\mathrm{l},*} = \mathcal{T}^{\mathrm{l}}_{Q^{\mathrm{h},*}} Q^{\mathrm{l},*}$.

We now want to show that the solution found by the Feudal Q-learning updates is a Stackelberg equilibrium, where the high-level is the leader and the low-level is the follower. In order to show this, we need to show that:

1. $(Q^{\mathrm{h},*}, Q^{\mathrm{l},*}) \in D^{\mathrm{l}}$;
2. $Q^{\mathrm{h}}_{Q^{\mathrm{h},*},Q^{\mathrm{l},*}} \geq Q^{\mathrm{h}}_{Q^{\mathrm{h}},Q^{\mathrm{l}}}, \quad \forall (Q^{\mathrm{h}}, Q^{\mathrm{l}}) \in D^{\mathrm{l}}$.

Point 1. is satisfied using the result stated in the paragraph on the Nash equilibrium, since $(Q^{\mathrm{h},*}, Q^{\mathrm{l},*}) \in D^{\mathrm{h}} \cap D^{\mathrm{l}}$ implies that $(Q^{\mathrm{h},*}, Q^{\mathrm{l},*}) \in D^{\mathrm{l}}$.

Point 2. can be shown by contradiction. Assume that there exist one or more pairs $(Q^{\mathrm{h}}, Q^{\mathrm{l}}) \in D^{\mathrm{l}}$ such that $Q^{\mathrm{h}}_{Q^{\mathrm{h}},Q^{\mathrm{l}}} > Q^{\mathrm{h}}_{Q^{\mathrm{h},*},Q^{\mathrm{l},*}}$ and $(Q^{\mathrm{h}}, Q^{\mathrm{l}}) \neq (Q^{\mathrm{h},*}, Q^{\mathrm{l},*})$. Let us call $C$ the non-empty set of such pairs.

Let us consider the pair $(\bar{Q}^{\mathrm{h}}, \bar{Q}^{\mathrm{l}})$ that maximizes $Q^{\mathrm{h}}_{Q^{\mathrm{h}},Q^{\mathrm{l}}}$, namely, $Q^{\mathrm{h}}_{\bar{Q}^{\mathrm{h}},\bar{Q}^{\mathrm{l}}} \geq Q^{\mathrm{h}}_{Q^{\mathrm{h}},Q^{\mathrm{l}}}\ \forall (Q^{\mathrm{h}}, Q^{\mathrm{l}}) \in C$.

The condition $(\bar{Q}^{\mathrm{h}}, \bar{Q}^{\mathrm{l}}) \in D^{\mathrm{l}}$ implies that Eq. (7b) is satisfied, i.e., finding the unique fixed point that satisfies $\bar{Q}^{\mathrm{l}} = \mathcal{T}^{\mathrm{l}}_{\bar{Q}^{\mathrm{h}}} \bar{Q}^{\mathrm{l}}$.

For a given low-level Q-function, $Q^{\mathrm{l}}$, the optimal high-level Q-function is unique and it is the only fixed point of the operator $\mathcal{T}^{\mathrm{h}}_{Q^{\mathrm{l}}}$. Thus, $\bar{Q}^{\mathrm{h}} = \mathcal{T}^{\mathrm{h}}_{\bar{Q}^{\mathrm{l}}} \bar{Q}^{\mathrm{h}}$. Thus we have found the unique pair $(\bar{Q}^{\mathrm{h}}, \bar{Q}^{\mathrm{l}})$ satisfying $(\bar{Q}^{\mathrm{h}}, \bar{Q}^{\mathrm{l}}) \in D^{\mathrm{h}} \cap D^{\mathrm{l}}$. Since it is the unique pair satisfying $(\bar{Q}^{\mathrm{h}}, \bar{Q}^{\mathrm{l}}) \in D^{\mathrm{h}} \cap D^{\mathrm{l}}$ and since $(Q^{\mathrm{h},*}, Q^{\mathrm{l},*})$ satisfies the condition, $Q^{\mathrm{h}}_{\bar{Q}^{\mathrm{h}},\bar{Q}^{\mathrm{l}}} = Q^{\mathrm{h}}_{Q^{\mathrm{h},*},Q^{\mathrm{l},*}}$ and $(\bar{Q}^{\mathrm{h}}, \bar{Q}^{\mathrm{l}}) = (Q^{\mathrm{h},*}, Q^{\mathrm{l},*})$, reaching a contradiction.

Moreover, the other pairs $(Q^{\mathrm{h}}, Q^{\mathrm{l}})$ in $C$ either lead to $Q^{\mathrm{h}}_{\bar{Q}^{\mathrm{h}},\bar{Q}^{\mathrm{l}}} = Q^{\mathrm{h}}_{Q^{\mathrm{h}},Q^{\mathrm{l}}}$ or to $Q^{\mathrm{h}}_{\bar{Q}^{\mathrm{h}},\bar{Q}^{\mathrm{l}}} > Q^{\mathrm{h}}_{Q^{\mathrm{h}},Q^{\mathrm{l}}}$. On the one hand, the equality case can be ruled out by arguments similar to those used for the pair $(\bar{Q}^{\mathrm{h}}, \bar{Q}^{\mathrm{l}})$. On the other hand, the inequality case would result in $(Q^{\mathrm{h}}, Q^{\mathrm{l}})$ pairs for which $Q^{\mathrm{h}}_{Q^{\mathrm{h}},Q^{\mathrm{l}}} < Q^{\mathrm{h}}_{Q^{\mathrm{h},*},Q^{\mathrm{l},*}}$.

Thus, there is no other pair $(Q^{\mathrm{h}}, Q^{\mathrm{l}}) \in D^{\mathrm{l}}$ such that $Q^{\mathrm{h}}_{Q^{\mathrm{h}},Q^{\mathrm{l}}} > Q^{\mathrm{h}}_{Q^{\mathrm{h},*},Q^{\mathrm{l},*}}$. □

## 5. Numerical experiments

Based on the proposed Feudal Q-learning updates in Eqs. (8a) and (8b), we test our HRL algorithm in a Minigrid environment [46] shown in Fig. 3. The goal is to make the agent, always starting in the same initial state $s_0$ (the light blue triangle in Fig. 3 represents the initial pose), reach a specific position on the map (green tile). At each time step, the agent can either turn 90° clockwise, 90° counterclockwise, or move forward. The high-level policy chooses tiles from a subset of $\mathcal{S}$ as the goal for the low-level policy, and it receives a positive reward when the agent moves from a black tile to the green tile; otherwise, it gets no reward. The low-level policy receives a positive reward when the agent has reached the goal position set by the high-level policy. After a predefined number of time steps, the episode ends.

In order to satisfy Assumption 4, we parameterize the step-sizes in Eq. (8) as follows:

$$\alpha(n) = \left(\frac{\alpha_0}{\sqrt[c]{\tilde{N}^c + n^c}}\right)^{E_\alpha}, \quad \beta(n) = \left(\frac{\beta_0}{\sqrt[c]{\tilde{N}^c + n^c}}\right)^{E_\beta}, \tag{24}$$

with $\tilde{N} \geq 0$, $\frac{1}{2} < E_\alpha < E_\beta \leq 1$, $\alpha_0$, $\beta_0$, $c > 0$ as free parameters. In the experiments we choose $\tilde{N} = 5000$, $E_\alpha = 0.6$, $E_\beta = 0.8$, $\alpha_0 \approx 1500$, $\beta_0 \approx 50$, $c = 5$ unless stated otherwise.

We design three experiments, reported in Sections 5.1–5.3, that respectively consider the following questions:

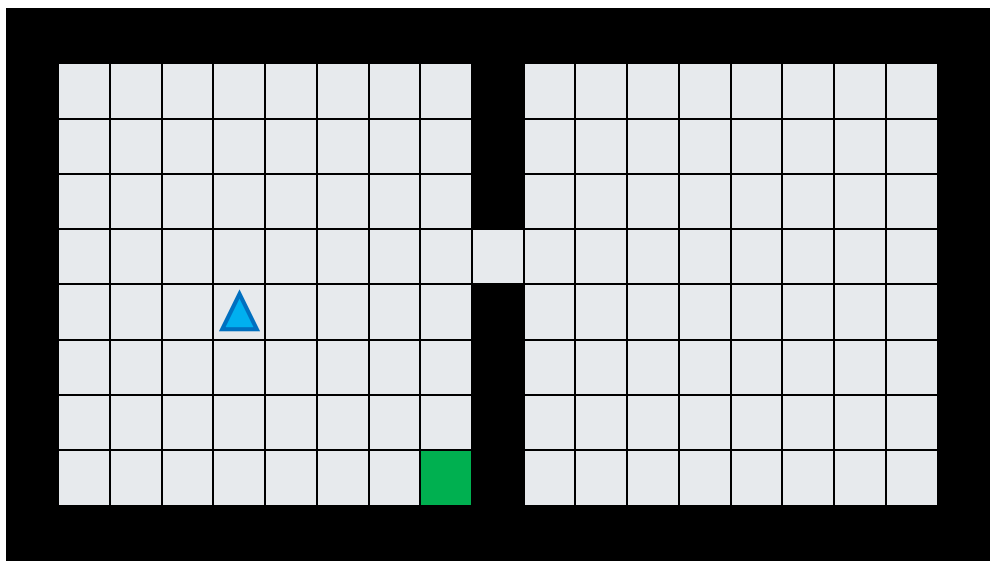

**Fig. 3.** Representation of the grid environment. The light blue triangle represents the initial pose of the agent, while the green square is the tile to be reached. Black tiles represent walls.

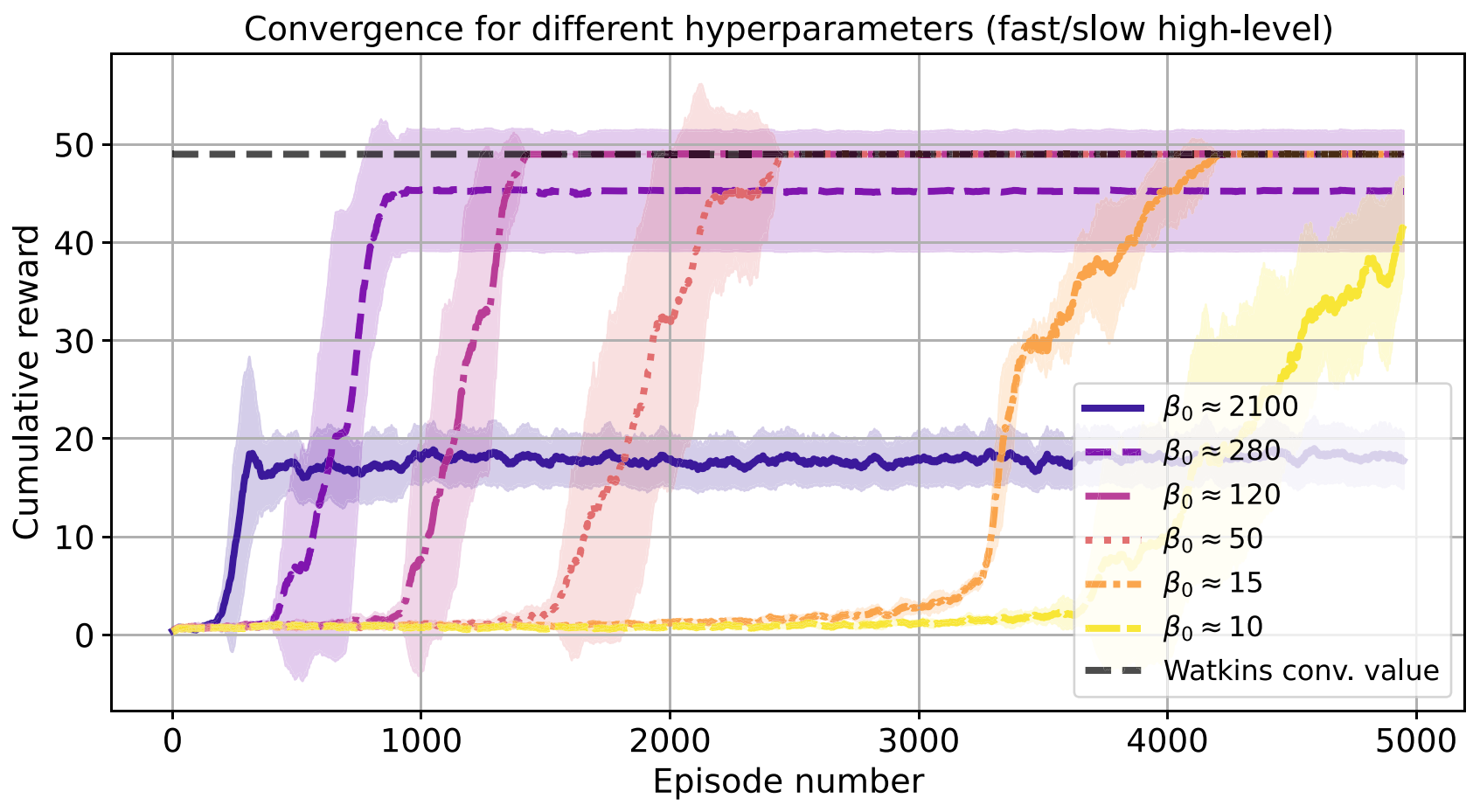

**Fig. 4.** The experiment shows the convergent behaviour predicted by Theorem 1. Moreover, the HRL agent is converging to the same cumulative reward as the Watkins' Q-learning agent. A faster high-level (bigger $\beta_0$) makes the algorithm converge faster, but being too fast can hinder convergence to the best policy.

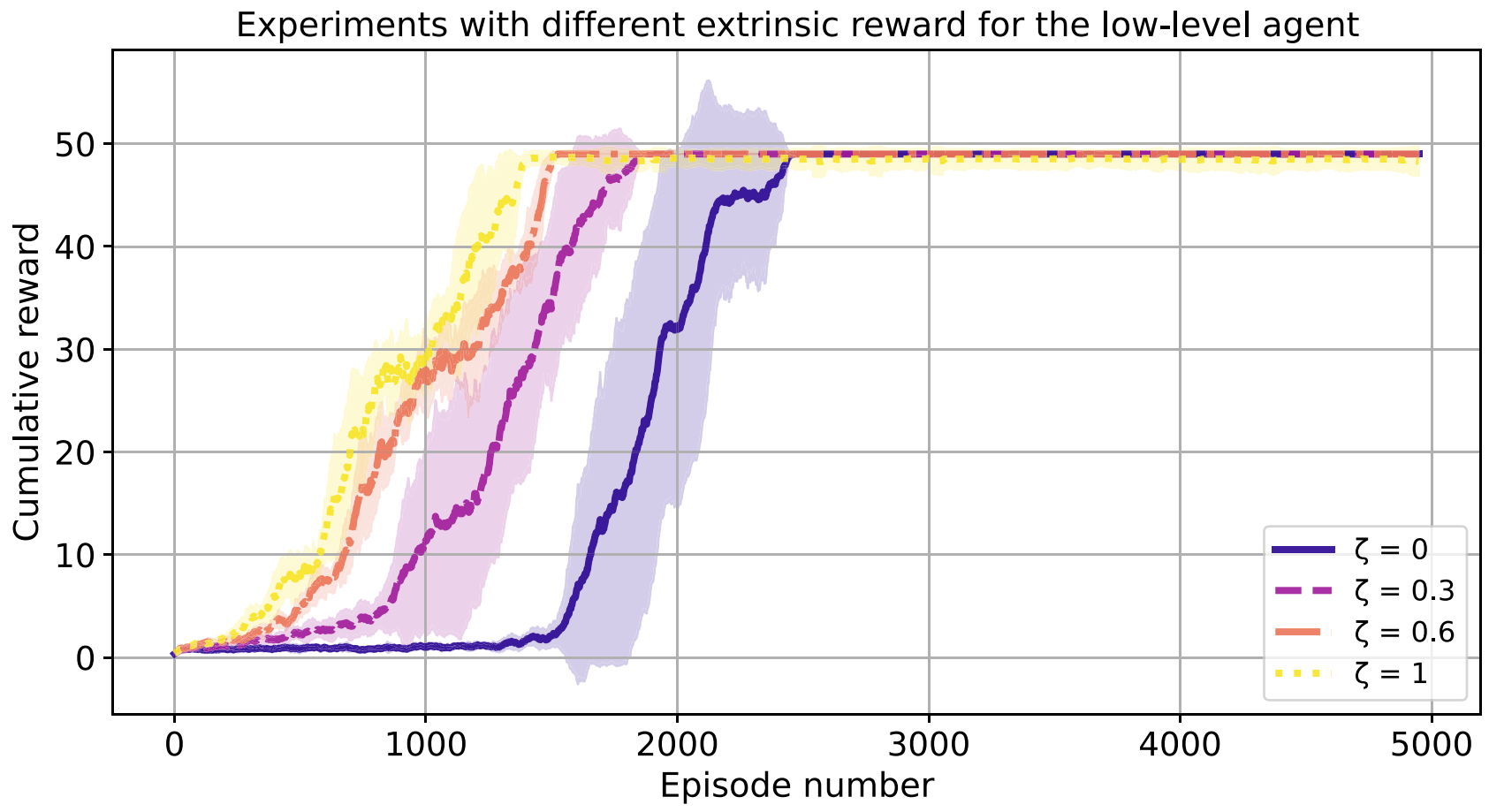

**Fig. 5.** An experimental study on the introduction of extrinsic reward for the low-level policy. A higher $\zeta$ corresponds to weighting more the extrinsic reward for the low-level policy. Introducing some extrinsic reward for the low-level policy is beneficial for faster convergence of the learning algorithm.

1) How does the Feudal Q-learning algorithm converge in practice? Does it learn a policy that accumulates as much reward as a standard Q-learning policy? What happens when the timescale separation is made more or less prominent by changing the high- and low-level step-sizes?
2) What is the influence of the extrinsic reward signal at the low-level? Does it speed up convergence?
3) Does the Feudal Q-learning algorithm help with transfer learning?

### 5.1. Convergence of Feudal Q-learning

In the first experiment, we train the HRL agent for 5000 episodes, where the green tile is positioned as in Fig. 3. An episode consists of the agent interacting with the environment for 300 low-level time steps. After $\approx$ 886 episodes, the agent has learnt a high-level policy and a low-level one that enable it to achieve a cumulative reward of 49 consistently. We plot in Fig. 4 the mean value and the standard deviation of the cumulative reward per episode of HRL training runs with different hyperparameters. The plots are smoothed using a moving average that considers 50 episodes. The mean and standard deviation are computed over 5 simulations.[1]

Across the different trainings, the low-level parameters are kept fixed, while we make the high-level algorithm faster or slower. In order to do so, the parameter $\beta_0$ in Eq. (24) is varied, changing the high-level step-size. A higher $\beta_0$ corresponds to faster high-level learning. We clearly see in Fig. 4 that faster high-level learning makes the algorithm converge faster, but, in line with the result and proof strategy of Theorem 1, overly fast high-level learning hinders convergence to the optimal value. Additionally, in Fig. 4, we also plot the cumulative reward value to which the Watkins' Q-learning algorithm converges.

While we have analyzed here the effect of tuning one hyperparameter, namely $\beta_0$, we highlight that the hyperparameters of the experiments have not been optimized. Thus, we foresee that better performance, e.g., faster convergence, can be obtained by properly tuning the hyperparameters. A study of the algorithm's transient performance is out of the scope of the paper.

### 5.2. Extrinsic reward at the low-level

In the second experiment, we consider the effect of the extrinsic reward in the low-level infinite-horizon problem, which has been shown to be beneficial in practice [13]. We want to see if those benefits also apply to our architecture.

In these experiments, in addition to the intrinsic reward, the low-level policy receives the same extrinsic reward $r$ that the high-level policy receives from the environment, weighted by a coefficient $\zeta \geq 0$. Thus, when setting $\zeta = 0$, the low-level policy does not receive any extrinsic reward. In Fig. 5, the results of different trainings with different $\zeta$ values are reported. One can easily see how introducing some extrinsic reward ($\zeta > 0$) speeds up convergence. Thus, incorporating an extrinsic reward in the low-level infinite-horizon problem can accelerate learning in our architecture, consistent with the motivation discussed in Section 1.

[1] On a CPU (AMD EPYC 7702), one simulation takes on average 36 min.

### 5.3. Transfer learning capabilities

In the third experiment we consider a more challenging scenario: the green tile is moved to a different room, as shown in Fig. 6. We investigate whether pretraining an HRL agent on a first goal helps learn a second goal, or whether it is counterproductive compared to training an HRL agent from scratch on the second goal. On the one hand, an HRL agent pretrained on the first goal could facilitate learning the second goal because the low-level policy may have acquired reusable goal-reaching skills. On the other hand, learning to reach the first goal could bias the agent toward that goal and thereby impede learning to reach a different goal.

The first agent that we consider is an HRL agent that, after being trained on the first goal as in Section 5.1 (Fig. 3), is retrained for 10,000 episodes on the second goal (Fig. 6). In the retraining phase, the low-level policy is greedy, while the high-level policy has the freedom to explore. The second agent that we consider is an HRL agent trained from scratch for 10,000 episodes on the second goal (Fig. 6), similar to what is done in Section 5.1.

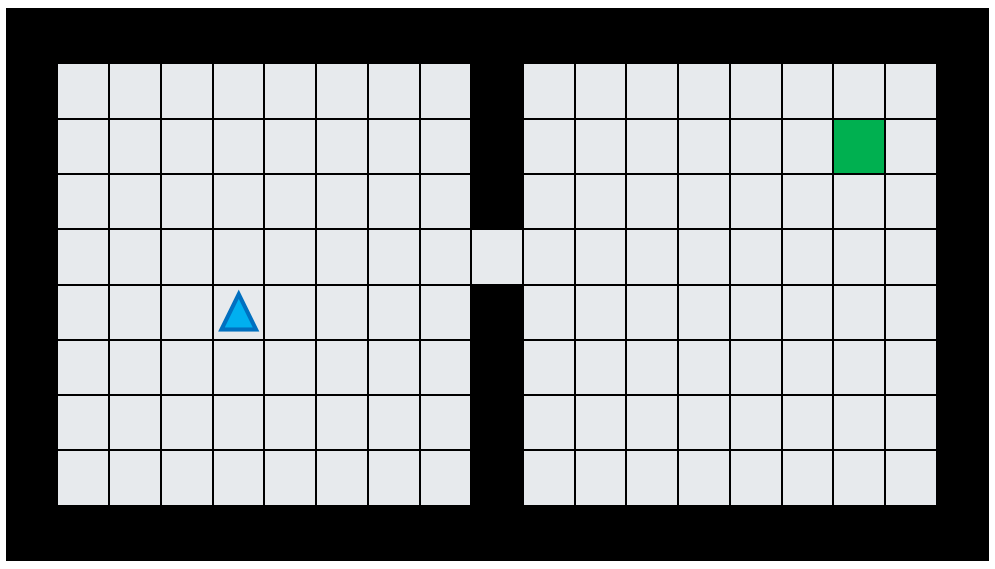

**Fig. 6.** Representation of the grid environment for the second goal. The light blue triangle represents the initial pose of the agent, while the green square is the tile to be reached. Black tiles represent walls.

We plot in Fig. 7 the mean value and the standard deviation of the cumulative reward per episode of the first agent (pretrained) and of the second agent (directly trained on the second goal). The pretrained agent converges in $\approx 6816$ episodes, while the second agent converges in $\approx 8620$ episodes. We see how Feudal Q-learning helps with transfer learning: retraining the agent with a different goal requires $\approx 20.93\%$ fewer episodes for convergence to a better cumulative reward. Moreover, the pretrained agent shows a smaller variance during training. This likely reflects the fact that the low-level policy retains useful goal-tracking behaviour, easing adaptation.

## 6. Conclusions and outlook

We presented an HRL scheme, termed Feudal Q-learning, and theoretical results stating its convergence and stability. To the best of the authors' knowledge, this is the first work guaranteeing the stability of an HRL algorithm.

The non-stationarity problems typical of HRL are directly tackled using a timescale separation stemming from the theory of singularly perturbed ODEs. The dynamical systems interpretation of the algorithm enabled a general analysis that delivered new theoretical insights on sufficient conditions for achieving desired properties of the algorithm. Future work could include analyzing HRL algorithms with other system theoretic techniques, e.g., the small gain theorem, or designing alternative HRL algorithms by employing the ODE method as described in [39, Chapter 8].

Additionally, the proposed scheme naturally connects to the field of game theory: we have multiple agents interacting, $\pi^{\mathrm{h}}$ and $\pi^{\mathrm{l}}$, while having potentially different goals, i.e., respectively maximizing $Q^{\mathrm{h}}_{\pi^{\mathrm{h}},\pi^{\mathrm{l}}}$ and $Q^{\mathrm{l}}_{\pi^{\mathrm{h}},\pi^{\mathrm{l}}}$. Future work should further explore the connections between HRL and game theory.

## CRediT authorship contribution statement

**Massimiliano Manenti:** Writing – review & editing, Writing – original draft, Visualization, Validation, Software, Resources, Project administration, Methodology, Investigation, Formal analysis, Data curation, Conceptualization. **Andrea Iannelli:** Writing – review & editing, Validation, Supervision, Resources, Project administration, Methodology, Funding acquisition, Conceptualization.

## Declaration of Generative AI and AI-assisted technologies in the writing process

During the preparation of this work the author(s) used ChatGPT to check for grammatical mistakes and typos. After using this tool/service,

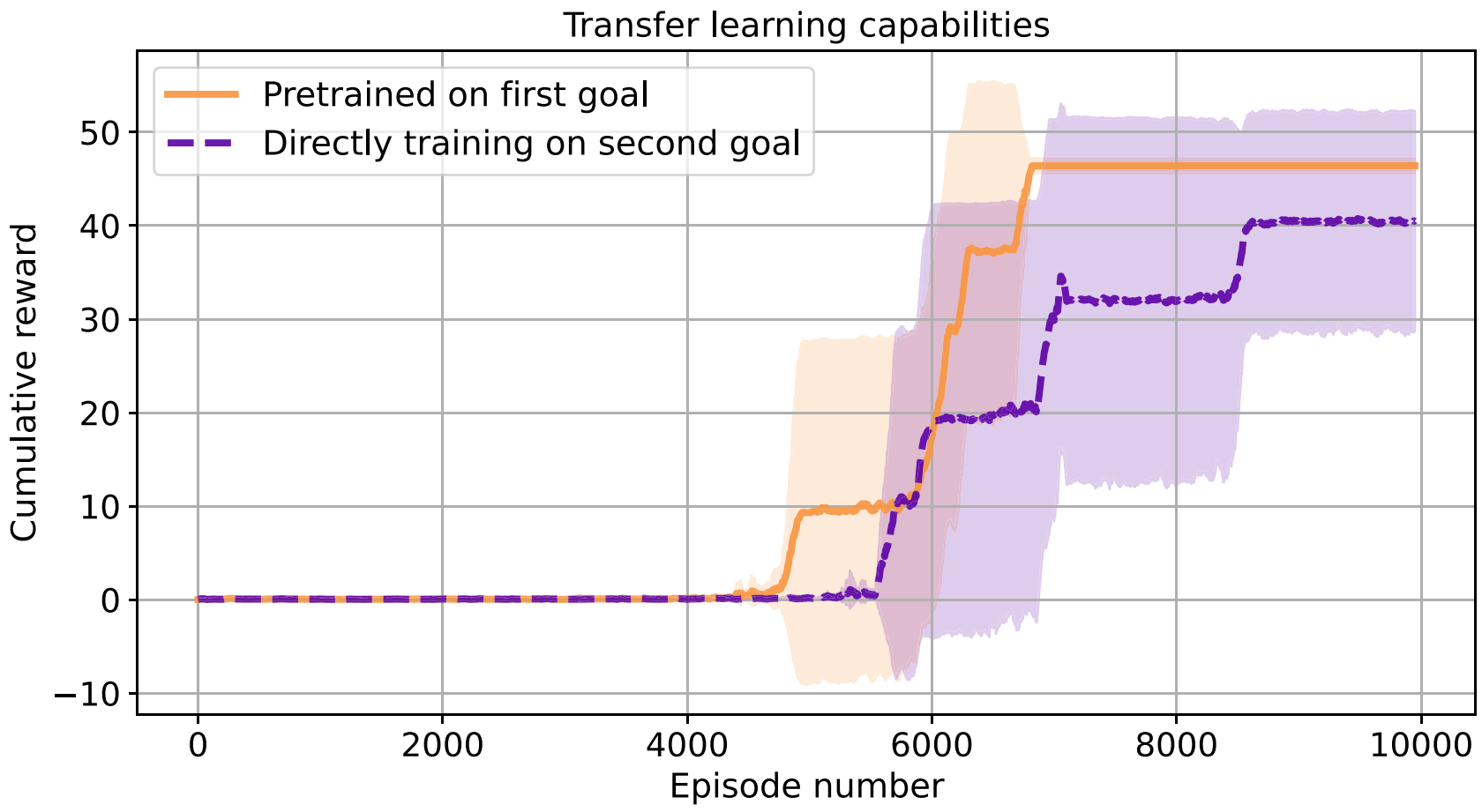

**Fig. 7.** Experiment showing the transfer learning capabilities of the HRL agent. The agent pretrained on the first goal (continuous orange line) converges faster and achieves a better cumulative reward than the agent trained from scratch on the second goal (dashed purple line).

the author(s) reviewed and edited the content as needed and take(s) full responsibility for the content of the published article.

## Declaration of competing interest

The authors declare that they have no known competing financial interests or personal relationships that could have appeared to influence the work reported in this paper.

## Acknowledgements

Founded by Deutsche Forschungsgemeinschaft (DFG, German Research Foundation) under Germany's Excellence Strategy - EXC 2075–390740016. The authors acknowledge the support of the Stuttgart Center for Simulation Science (SimTech).

## Appendix A. Details on the formulation of Feudal Q-learning as TTSA iterates

As reported in Section 3.2, we rewrite the Feudal Q-learning updates in Eq. (8) as Eq. (10), in a form consistent with the TTSA literature's notation [25,38]. Being in a finite sets setting, we treat $Q^{\mathrm{h},n}$ and $Q^{\mathrm{l},n}$ as matrices and define $x_n$ and $y_n$ as the vectorized versions of $Q^{\mathrm{l},n}$ and $Q^{\mathrm{h},n}$, respectively. In particular, for the high-level:

- $y_n := \mathrm{vec}(Q^{\mathrm{h},n})$.
- We define $\rho^{\mathrm{h}}_{s^{\mathrm{h}}} \in \{1,\dots,|S^{\mathrm{h}}|\}$ as the row number that state $s^{\mathrm{h}}$ occupies in $Q^{\mathrm{h},n}$. Analogously, we define $c^{\mathrm{h}}_{\omega^{\mathrm{h}}} \in \{1,\dots,|\Omega|\}$ as the column number that goal $\omega^{\mathrm{h}}$ occupies in $Q^{\mathrm{h},n}$. $g(x_n,y_n) : \mathbb{R}^{|S^{\mathrm{l}}|\times|\mathcal{A}|} \times \mathbb{R}^{|S^{\mathrm{h}}|\times|\Omega|} \to \mathbb{R}^{|S^{\mathrm{h}}|\times|\Omega|}$ is the vector-valued function whose $((c^{\mathrm{h}}_{\omega^{\mathrm{h}}}-1)|S^{\mathrm{h}}|+\rho^{\mathrm{h}}_{s^{\mathrm{h}}})$-th component is $g_{s^{\mathrm{h}},\omega^{\mathrm{h}}}(x_n,y_n)$, where: $g_{s^{\mathrm{h}},\omega^{\mathrm{h}}}(x_n,y_n) = r^{\mathrm{h}}_{\pi^{\mathrm{l}}}(s^{\mathrm{h}},\omega^{\mathrm{h}}) + (\gamma^{\mathrm{h}})^T\, \mathbb{E}_{P^{\mathrm{h}}_{\pi^{\mathrm{l}}}}[\max_{\omega^+} Q^{\mathrm{h},n}(s^{\mathrm{h},+},\omega^+)] - Q^{\mathrm{h},n}(s^{\mathrm{h}},\omega^{\mathrm{h}})$.
- $M^{(2)}_{n+1}(x_n,y_n)$ is the vector-valued function whose $((c^{\mathrm{h}}_{\omega^{\mathrm{h}}}-1)|S^{\mathrm{h}}|+\rho^{\mathrm{h}}_{s^{\mathrm{h}}})$-th component is $M^{(2)}_{n+1,s^{\mathrm{h}},\omega^{\mathrm{h}}}(x_n,y_n)$.
- $M^{(2)}_{n+1,s^{\mathrm{h}},\omega^{\mathrm{h}}}(x_n,y_n) := (\gamma^{\mathrm{h}})^T \max_{\omega} Q^{\mathrm{h},n}(s^{\mathrm{h}}_{t+1},\omega) - (\gamma^{\mathrm{h}})^T\, \mathbb{E}_{s^{\mathrm{h},+}\sim P^{\mathrm{h}}_{\pi^{\mathrm{l}}}}[\max_{\omega^+} Q^{\mathrm{h},n}(s^{\mathrm{h},+},\omega^+)] + \sum_{k=0}^{T-1}(\gamma^{\mathrm{h}})^k r(s_{tT+k},a_{tT+k}) - r^{\mathrm{h}}_{\pi^{\mathrm{l}}}(s^{\mathrm{h}},\omega^{\mathrm{h}})$, where $s^{\mathrm{h}}_{t+1} \sim P^{\mathrm{h}}_{\pi^{\mathrm{l}}}(\cdot|s^{\mathrm{h}},\omega^{\mathrm{h}})$, $a_{tT+k} \sim \pi^{\mathrm{l}}(\cdot|s^{\mathrm{l}}_{tT+k})$, and $s^{\mathrm{l}}_{tT+k+1} \sim P^{\mathrm{l}}_{\pi^{\mathrm{h}}}(\cdot|s^{\mathrm{l}}_{tT+k},a_{tT+k})$.

Whereas, for the low-level:

- $x_n := \mathrm{vec}(Q^{\mathrm{l},n})$.
- We define $\rho^{\mathrm{l}}_{s^{\mathrm{l}}} \in \{1,\dots,|S^{\mathrm{l}}|\}$ as the row number that state $s^{\mathrm{l}}$ occupies in $Q^{\mathrm{l},n}$. Analogously, we define $c^{\mathrm{l}}_a \in \{1,\dots,|\mathcal{A}|\}$ as the column number that action $a$ occupies in $Q^{\mathrm{l},n}$. $h(x_n,y_n) : \mathbb{R}^{|S^{\mathrm{l}}|\times|\mathcal{A}|}\times\mathbb{R}^{|S^{\mathrm{h}}|\times|\Omega|} \to \mathbb{R}^{|S^{\mathrm{l}}|\times|\mathcal{A}|}$ is the vector-valued function whose $((c^{\mathrm{l}}_a-1)|S^{\mathrm{l}}|+\rho^{\mathrm{l}}_{s^{\mathrm{l}}})$-th component is $h_{s^{\mathrm{l}},a}(x_n,y_n)$, where $h_{s^{\mathrm{l}},a}(x_n,y_n) = r^{\mathrm{l}}(s^{\mathrm{l}},a) + \gamma^{\mathrm{l}}\, \mathbb{E}_{P^{\mathrm{l}}_{\pi^{\mathrm{h}}}}[\max_{a^+} Q^{\mathrm{l},n}(s^{\mathrm{l},+},a^+)] - Q^{\mathrm{l},n}(s^{\mathrm{l}},a)$.
- $M^{(1)}_{n+1}(x_n,y_n)$ is the vector-valued function whose $((c^{\mathrm{l}}_a-1)|S^{\mathrm{l}}|+\rho^{\mathrm{l}}_{s^{\mathrm{l}}})$-th component is $M^{(1)}_{n+1,s^{\mathrm{l}},a}(x_n,y_n)$.
- $M^{(1)}_{n+1,s^{\mathrm{l}},a}(x_n,y_n) := \gamma^{\mathrm{l}} \max_a Q^{\mathrm{l},n}(s^{\mathrm{l}}_{k+1},a) - \gamma^{\mathrm{l}}\, \mathbb{E}_{s^{\mathrm{l},+}\sim P^{\mathrm{l}}_{\pi^{\mathrm{h}}}}[\max_{a^+} Q^{\mathrm{l},n}(s^{\mathrm{l},+},a^+)]$, where $s^{\mathrm{l}}_{k+1} \sim P^{\mathrm{l}}_{\pi^{\mathrm{h}}}(\cdot|s^{\mathrm{l}},a)$.

In this way, we have shown how to rewrite the Feudal Q-learning updates in Eq. (8) as the TTSA updates in Eq. (10).

## Appendix B. Derivation of high- and low-level dynamics

For ease of exposition, we will use $P(\cdot)$ to indicate a probability that is clear from the context. A high-level trajectory between two time steps $t_1 \le t_2$ is $\tau^{\mathrm{h}}_{[t_1,t_2]} := (s^{\mathrm{h}}_{t_1},\omega^{\mathrm{h}}_{t_1},s^{\mathrm{h}}_{t_1+1},\omega^{\mathrm{h}}_{t_1+1},\dots,s^{\mathrm{h}}_{t_2},\omega^{\mathrm{h}}_{t_2})$. Moreover, we indicate with $\tau^{\mathrm{h}}_t$ a high-level trajectory starting from the initial time, i.e., $\tau^{\mathrm{h}}_t := \tau^{\mathrm{h}}_{[0,t]}$. Similarly, for the low-level MDP we define $\tau^{\mathrm{l}}_{[k_1,k_2]} := (s^{\mathrm{l}}_{k_1},a_{k_1},\dots,s^{\mathrm{l}}_{k_2},a_{k_2})$ and $\tau^{\mathrm{l}}_k := \tau^{\mathrm{l}}_{[0,k]}$.

### B.1. High-level dynamics

Since the evolution of $s^{\mathrm{de}}$ between two consecutive decision epochs is known and deterministic, and since the goal $\omega^{\mathrm{l}}_k$ is constant between decision epochs, the probability $P^{\mathrm{h}}_{\pi^{\mathrm{l}}}(s^{\mathrm{h}}_1|s^{\mathrm{h}}_0,\omega^{\mathrm{h}}_0)$ is equal to $P(s_T|s_0,\omega^{\mathrm{l}}_0 = \dots = \omega^{\mathrm{l}}_{T-1} = \omega^{\mathrm{h}}_t, s^{\mathrm{de}}_0 = 0,\dots,s^{\mathrm{de}}_{T-1} = T-1)$. We will denote by $E$ the event where $\omega^{\mathrm{l}}_0 = \dots = \omega^{\mathrm{l}}_{T-1} = \omega^{\mathrm{h}}_t$ and $s^{\mathrm{de}}_0 = 0,\dots s^{\mathrm{de}}_{T-1} = T-1$. Then, $P(s_T|s_0,E)$ can be written as:

$$
\begin{aligned}
&P(s_T|s_0,E)\\
&= \sum_{\substack{s_1,\dots,s_{T-1}\in S\\ a_0,\dots,a_{T-1}\in\mathcal{A}}} P(s_T|s_0,E,a_0,s_1,a_1,\dots,s_{T-1},a_{T-1})P(a_0,s_1,a_1,\dots,s_{T-1},a_{T-1}|s_0,E)\\
&= \sum_{\substack{s_1,\dots,s_{T-1}\in S\\ a_0,\dots,a_{T-1}\in\mathcal{A}}} P\left(s_T|\tau^{\mathrm{l}}_{T-1}\right) P\left(\tau^{\mathrm{l}}_{T-1}|s_0,E\right)\\
&= \sum_{\substack{s_1,\dots,s_{T-1}\in S\\ a_0,\dots,a_{T-1}\in\mathcal{A}}} P(s_T|s_{T-1},a_{T-1}) P\left(\tau^{\mathrm{l}}_{T-1}|s_0,E\right)\\
&= \sum_{\substack{s_1,\dots,s_{T-1}\in S\\ a_0,\dots,a_{T-1}\in\mathcal{A}}} P(s_T|s_{T-1},a_{T-1}) P\left(a_{T-1}|s_{T-1},\tau^{\mathrm{l}}_{T-2}\right) P\left(s_{T-1},\tau^{\mathrm{l}}_{T-2}|s_0,E\right)\\
&= \sum_{\substack{s_1,\dots,s_{T-1}\in S\\ a_0,\dots,a_{T-1}\in\mathcal{A}}} P(s_T|s_{T-1},a_{T-1}) \pi^{\mathrm{l}}\left(a_{T-1}|s^{\mathrm{l}}_{T-1}\right) P\left(s_{T-1},\tau^{\mathrm{l}}_{T-2}|s_0,E\right)\\
&= \sum_{\substack{s_1,\dots,s_{T-1}\in S\\ a_0,\dots,a_{T-1}\in\mathcal{A}}} P(s_T|s_{T-1},a_{T-1}) \pi^{\mathrm{l}}\left(a_{T-1}|s^{\mathrm{l}}_{T-1}\right) P(s_{T-1}|s_{T-2},a_{T-2}) P\left(\tau^{\mathrm{l}}_{T-2}|s_0,E\right)\\
&= \cdots \text{ (iteratively perform steps above)}\\
&= \sum_{\substack{s_1,\dots,s_{T-1}\in S\\ a_0,\dots,a_{T-1}\in\mathcal{A}}} \prod_{i=1}^{T} P(s_i|s_{i-1},a_{i-1}) \pi^{\mathrm{l}}\left(a_{i-1}|s_{i-1},\omega^{\mathrm{l}}_{i-1},s^{\mathrm{de}}_{i-1}\right).
\end{aligned}
$$

where the first equality follows from the law of total probability, the third from the Markov property, the fourth from the chain rule, and the fifth from combining the Markov property with the chain rule.

We have shown how to derive $P^{\mathrm{h}}_{\pi^{\mathrm{l}}}(s^{\mathrm{h}}_1|s^{\mathrm{h}}_0,\omega^{\mathrm{h}}_0)$ for the sake of exposition, similar steps readily apply to derive $P^{\mathrm{h}}_{\pi^{\mathrm{l}}}(s^{\mathrm{h}}_{t+1}|s^{\mathrm{h}}_t,\omega^{\mathrm{h}}_t)$.

### B.2. Low-level dynamics

By applying the chain rule (first and third equality in Eq. B.1) and properly managing independent variables (second and fourth equality in Eq. B.1), we can write the low-level dynamics as:

$$
\begin{aligned}
&P^{\mathrm{l}}_{\pi^{\mathrm{h}}}\left(s_{k+1},\omega^{\mathrm{l}}_{k+1},s^{\mathrm{de}}_{k+1}|s_k,\omega^{\mathrm{l}}_k,s^{\mathrm{de}}_k,a_k\right)\\
&= P\left(s^{\mathrm{de}}_{k+1}|s_{k+1},\omega^{\mathrm{l}}_{k+1},s_k,\omega^{\mathrm{l}}_k,s^{\mathrm{de}}_k,a_k\right) P\left(s_{k+1},\omega^{\mathrm{l}}_{k+1}|s_k,\omega^{\mathrm{l}}_k,s^{\mathrm{de}}_k,a_k\right)\\
&= P\left(s^{\mathrm{de}}_{k+1}|s^{\mathrm{de}}_k\right) P\left(s_{k+1},\omega^{\mathrm{l}}_{k+1}|s_k,\omega^{\mathrm{l}}_k,s^{\mathrm{de}}_k,a_k\right)\\
&= P\left(s^{\mathrm{de}}_{k+1}|s^{\mathrm{de}}_k\right) P\left(s_{k+1}|\omega^{\mathrm{l}}_{k+1},s_k,\omega^{\mathrm{l}}_k,s^{\mathrm{de}}_k,a_k\right) P\left(\omega^{\mathrm{l}}_{k+1}|s_k,\omega^{\mathrm{l}}_k,s^{\mathrm{de}}_k,a_k\right)\\
&= P\left(s^{\mathrm{de}}_{k+1}|s^{\mathrm{de}}_k\right) P(s_{k+1}|s_k,a_k) P\left(\omega^{\mathrm{l}}_{k+1}|s_k,\omega^{\mathrm{l}}_k,s^{\mathrm{de}}_k\right).
\end{aligned} \tag{B.1}
$$

Knowing that the goal does not change unless we are at a decision epoch and that the high-level policy $\pi^{\mathrm{h}}$ selects goals, we have that:

$$
P\left(\omega^{\mathrm{l}}_{k+1}|s_k,\omega^{\mathrm{l}}_k,s^{\mathrm{de}}_k\right) = \begin{cases} 0, & \text{if } s^{\mathrm{de}}_k \neq T-1 \text{ and } \omega^{\mathrm{l}}_{k+1} \neq \omega^{\mathrm{l}}_k \\ 1, & \text{if } s^{\mathrm{de}}_k \neq T-1 \text{ and } \omega^{\mathrm{l}}_{k+1} = \omega^{\mathrm{l}}_k \\ \pi^{\mathrm{h}}\left(\omega^{\mathrm{l}}_{k+1}|s_k\right), & \text{if } s^{\mathrm{de}}_k = T-1 \end{cases}. \tag{B.2}
$$

We get the desired result by combining Eqs. (B.1) and (B.2).

## Appendix C. Proof of Theorem 1

We will provide lemmas that will be used to satisfy the assumptions needed to apply the TTSA results for stability (Theorem 10 in [38]) and convergence (Theorem 8.1 in [25]) of the iterates.

**Lemma 4 (Lipschitz continuity of $g$ and $h$).** *The functions g and h are Lipschitz in the infinity norm, i.e., there exist $L_g, L_h \geq 0$ such that, for any $(x^1, y^1), (x^2, y^2) \in \mathbb{R}^{|S^l|\times|\mathcal{A}|} \times \mathbb{R}^{|S^h|\times|\Omega|}$, it holds that:*

$$\left\|g(x^1, y^1) - g(x^2, y^2)\right\|_\infty \leq L_g \left\|(x^1, y^1) - (x^2, y^2)\right\|_\infty, \tag{C.1}$$

$$\left\|h(x^1, y^1) - h(x^2, y^2)\right\|_\infty \leq L_h \left\|(x^1, y^1) - (x^2, y^2)\right\|_\infty. \tag{C.2}$$

**Proof.** We will prove that $g$ is Lipschitz, with similar steps one can prove the Lipschitz continuity of $h$. We show that each component of $g$, i.e., $g_{s^h,\omega^h}$ is Lipschitz continuous, then the Lipschitz continuity of $g$ follows.

Recall that $g_{s^h,\omega^h}(Q^l, Q^h) = r^h_{\pi^l}(s^h, \omega^h) + (\gamma^h)^T \mathbb{E}_{s^{h,+}\sim P^h_{\pi^l}}[\max_{\omega^+} Q^h(s^{h,+}, \omega^+)] - Q^h(s^h, \omega^h)$.

To show the Lipschitz continuity of $g_{s^h,\omega^h}$, we will proceed by considering the three addends composing $g_{s^h,\omega^h}$. Then, since the sum of Lipschitz continuous functions is itself Lipschitz, $g_{s^h,\omega^h}$ is Lipschitz continuous with respect to $(Q^l, Q^h)$. The steps are listed in the following, and the more involved steps will be detailed below.

1. $r^h_{\pi^l}$ is Lipschitz continuous with respect to $(Q^l, Q^h)$:
   (a) We will show that $r^h_{\pi^l}$ is Lipschitz continuous with respect to $\pi^l$.
   (b) We will show that $\pi^l$ is Lipschitz continuous with respect to $Q^l$.
   (c) Since the composition of Lipschitz continuous functions is Lipschitz continuous, it is easy to show that $r^h_{\pi^l}(s^h, \omega^h)$ is Lipschitz continuous with respect to $(Q^l, Q^h)$.
2. $G^2_{Q^h,P^h_{\pi^l}} := (\gamma^h)^T \mathbb{E}_{s^{h,+}\sim P^h_{\pi^l}}[\max_{\omega^+} Q^h(s^{h,+}, \omega^+)]$ is Lipschitz continuous with respect to $(Q^l, Q^h)$:
   (a) We will show that $G^2_{Q^h,P^h_{\pi^l}}$ is Lipschitz continuous with respect to $Q^h$.
   (b) We will show that $G^2_{Q^h,P^h_{\pi^l}}$ is Lipschitz continuous with respect to $P^h_{\pi^l}$.
   (c) We will show that $P^h_{\pi^l}$ is Lipschitz continuous with respect to $\pi^l$.
   (d) The fact that $\pi^l$ is Lipschitz continuous with respect to $Q^l$ comes from 1b.
   (e) Since the composition of Lipschitz continuous functions is Lipschitz continuous, it is easy to show that $G^2_{Q^h,P^h_{\pi^l}}(s^h, \omega^h)$ is Lipschitz continuous with respect to $(Q^l, Q^h)$.
3. $-Q^h(s^h, \omega^h)$ is clearly Lipschitz continuous with respect to $(Q^l, Q^h)$.

*1a) Lipschitz continuity of $r^h_{\pi^l}$ with respect to $\pi^l$*

Firstly, the probability of a low-level trajectory, $P^{\tau^l_k}_{\pi^l,P^l}$, is:

$$P^{\tau^l_k}_{\pi^l,P^l} = \prod_{i=0}^{k} \pi^l\left(a_i|s^l_i\right) P^l_{\pi^h}\left(s^l_{i+1}|s^l_i, a_i\right),$$

and the high-level reward can be rewritten as:

$$r^h_{\pi^l}(s^h, \omega^h) = \sum_{k=0}^{T-1} (\gamma^h)^k \sum_{\tau^l_k} P^{\tau^l_k}_{\pi^l,P^l} r(s_k, a_k).$$

Thus:

$$\begin{aligned}
&|r^h_{\pi^{l,1}}(s^h, \omega^h) - r^h_{\pi^{l,2}}(s^h, \omega^h)| \\
&= \left|\sum_{k=0}^{T-1} (\gamma^h)^k \sum_{\tau^l_k} P^{\tau^l_k}_{\pi^{l,1},P^l} r(s_k, a_k) - \sum_{k=0}^{T-1} (\gamma^h)^k \sum_{\tau^l_k} P^{\tau^l_k}_{\pi^{l,2},P^l} r(s_k, a_k)\right| \\
&= \left|\sum_{k=0}^{T-1} (\gamma^h)^k \sum_{\tau^l_k} \left(P^{\tau^l_k}_{\pi^{l,1},P^l} - P^{\tau^l_k}_{\pi^{l,2},P^l}\right) r(s_k, a_k)\right| \\
&\leq \sum_{k=0}^{T-1} (\gamma^h)^k \sum_{\tau^l_k} \left|\left(P^{\tau^l_k}_{\pi^{l,1},P^l} - P^{\tau^l_k}_{\pi^{l,2},P^l}\right)\right| \left|r(s_k, a_k)\right| \\
&\leq \bar{r} \sum_{k=0}^{T-1} (\gamma^h)^k \sum_{\tau^l_k} \left|\left(P^{\tau^l_k}_{\pi^{l,1},P^l} - P^{\tau^l_k}_{\pi^{l,2},P^l}\right)\right|,
\end{aligned} \tag{C.3}$$

where the first inequality is a consequence of the triangle inequality, while the second uses the boundedness of the reward function assumed in Assumption 1. We will now find a bound on $|(P^{\tau^l_k}_{\pi^{l,1},P^l} - P^{\tau^l_k}_{\pi^{l,2},P^l})|$ that depends on $\left\|\pi^{l,1} - \pi^{l,2}\right\|_\infty$.

$$\begin{aligned}
&|\left( P^{\tau^l_k}_{\pi^{l,1},P^l} - P^{\tau^l_k}_{\pi^{l,2},P^l}\right)| \\
&= \left|\prod_{i=0}^{k} \pi^{l,1}\left(a_i|s^l_i\right) P^l_{\pi^h}\left(s^l_{i+1}|s^l_i, a_i\right) - \prod_{i=0}^{k} \pi^{l,2}\left(a_i|s^l_i\right) P^l_{\pi^h}\left(s^l_{i+1}|s^l_i, a_i\right)\right| \\
&= \left|\prod_{i=0}^{k} \pi^{l,1}\left(a_i|s^l_i\right) - \prod_{i=0}^{k} \pi^{l,2}\left(a_i|s^l_i\right)\right| \prod_{i=0}^{k} P^l_{\pi^h}\left(s^l_{i+1}|s^l_i, a_i\right) \\
&= \left|\sum_{j=0}^{k}\left(\prod_{i=0}^{j-1} \pi^{l,1}\left(a_i|s^l_i\right)\right)\left(\pi^{l,1}\left(a_j|s^l_j\right) - \pi^{l,2}\left(a_j|s^l_j\right)\right)\left(\prod_{i=j+1}^{k} \pi^{l,2}\left(a_i|s^l_i\right)\right)\right| \cdot \\
&\quad \cdot \prod_{i=0}^{k} P^l_{\pi^h}\left(s^l_{i+1}|s^l_i, a_i\right) \\
&\leq \sum_{j=0}^{k}\left|\left(\prod_{i=0}^{j-1} \pi^{l,1}\left(a_i|s^l_i\right)\right)\left(\pi^{l,1}\left(a_j|s^l_j\right) - \pi^{l,2}\left(a_j|s^l_j\right)\right)\left(\prod_{i=j+1}^{k} \pi^{l,2}\left(a_i|s^l_i\right)\right)\right| \cdot \\
&\quad \cdot \prod_{i=0}^{k} P^l_{\pi^h}\left(s^l_{i+1}|s^l_i, a_i\right) \\
&\leq \sum_{j=0}^{k}\left|\left(\pi^{l,1}\left(a_j|s^l_j\right) - \pi^{l,2}\left(a_j|s^l_j\right)\right)\right| \prod_{i=0}^{k} P^l_{\pi^h}\left(s^l_{i+1}|s^l_i, a_i\right) \\
&\leq (k+1)\left\|\pi^{l,1} - \pi^{l,2}\right\|_\infty \prod_{i=0}^{k} P^l_{\pi^h}\left(s^l_{i+1}|s^l_i, a_i\right),
\end{aligned} \tag{C.4}$$

where the third equality follows by expanding the difference of two products as a telescoping sum.

Now, we can insert (C.4) into (C.3):

$$\begin{aligned}
&|r^h_{\pi^{l,1}}(s^h, \omega^h) - r^h_{\pi^{l,2}}(s^h, \omega^h)| \\
&\leq \bar{r} \sum_{k=0}^{T-1} (\gamma^h)^k \sum_{\tau^l_k} \left|(k+1)\left\|\pi^{l,1} - \pi^{l,2}\right\|_\infty \prod_{i=0}^{k} P^l_{\pi^h}\left(s^l_{i+1}|s^l_i, a_i\right)\right| \\
&\leq \bar{r}\left\|\pi^{l,1} - \pi^{l,2}\right\|_\infty \sum_{k=0}^{T-1} (\gamma^h)^k (k+1) \sum_{\tau^l_k} \prod_{i=0}^{k} P^l_{\pi^h}\left(s^l_{i+1}|s^l_i, a_i\right).
\end{aligned} \tag{C.5}$$

Moreover, the term $\sum_{\tau^{\mathrm{l}}_k}\prod_{i=0}^{k} P^{\mathrm{l}}_{\pi^{\mathrm{h}}}(s^{\mathrm{l}}_{i+1}|s^{\mathrm{l}}_i,a_i)$ can be written as:

$$
\begin{aligned}
&\sum_{\tau^{\mathrm{l}}_k}\prod_{i=0}^{k} P^{\mathrm{l}}_{\pi^{\mathrm{h}}}\left(s^{\mathrm{l}}_{i+1}|s^{\mathrm{l}}_i,a_i\right)\\
&=\sum_{\substack{s^{\mathrm{l}}_0,\dots,s^{\mathrm{l}}_k\in\mathcal{S}^{\mathrm{l}}\\ a_0,\dots a^{\mathrm{l}}_k\in\mathcal{A}}}\prod_{i=0}^{k} P^{\mathrm{l}}_{\pi^{\mathrm{h}}}\left(s^{\mathrm{l}}_{i+1}|s^{\mathrm{l}}_i,a_i\right)\\
&=\sum_{s^{\mathrm{l}}_0}\sum_{a_0}\sum_{s^{\mathrm{l}}_1}P^{\mathrm{l}}_{\pi^{\mathrm{h}}}\left(s^{\mathrm{l}}_1|s^{\mathrm{l}}_0,a_0\right)\sum_{a_1}\sum_{s^{\mathrm{l}}_2}P^{\mathrm{l}}_{\pi^{\mathrm{h}}}\left(s^{\mathrm{l}}_2|s^{\mathrm{l}}_1,a_1\right)\cdots\sum_{a_{k-1}}\sum_{s^{\mathrm{l}}_k}P^{\mathrm{l}}_{\pi^{\mathrm{h}}}\left(s^{\mathrm{l}}_k|s^{\mathrm{l}}_{k-1},a_{k-1}\right)\\
&=\sum_{s^{\mathrm{l}}_0}\sum_{a_0}\sum_{s^{\mathrm{l}}_1}P^{\mathrm{l}}_{\pi^{\mathrm{h}}}\left(s^{\mathrm{l}}_1|s^{\mathrm{l}}_0,a_0\right)\sum_{a_1}\sum_{s^{\mathrm{l}}_2}P^{\mathrm{l}}_{\pi^{\mathrm{h}}}\left(s^{\mathrm{l}}_2|s^{\mathrm{l}}_1,a_1\right)\cdots\sum_{a_{k-1}}1\\
&=\sum_{s^{\mathrm{l}}_0}|\mathcal{A}|^k\\
&=|\mathcal{S}^{\mathrm{l}}||\mathcal{A}|^k. \qquad \text{(C.6)}
\end{aligned}
$$

Thus, inserting (C.6) in (C.5) and defining $L^{\pi^{\mathrm{l}}}_{r^{\mathrm{h}}} := \bar{r}\sum_{k=0}^{T-1}(\gamma^{\mathrm{h}})^k(k+1)|\mathcal{S}^{\mathrm{l}}||\mathcal{A}|^k$, we get:

$$
\begin{aligned}
|r^{\mathrm{h}}_{\pi^{\mathrm{l},1}}(s^{\mathrm{h}},\omega^{\mathrm{h}})-r^{\mathrm{h}}_{\pi^{\mathrm{l},2}}(s^{\mathrm{h}},\omega^{\mathrm{h}})| &\le \bar{r}\left\|\pi^{\mathrm{l},1}-\pi^{\mathrm{l},2}\right\|_\infty\sum_{k=0}^{T-1}(\gamma^{\mathrm{h}})^k(k+1)|\mathcal{S}^{\mathrm{l}}||\mathcal{A}|^k\\
&=L^{\pi^{\mathrm{l}}}_{r^{\mathrm{h}}}\left\|\pi^{\mathrm{l},1}-\pi^{\mathrm{l},2}\right\|_\infty, \qquad \text{(C.7)}
\end{aligned}
$$

which shows that $r^{\mathrm{h}}_{\pi^{\mathrm{l}}}$ is Lipschitz continuous with respect to $\pi^{\mathrm{l}}$.

*1b) Lipschitz continuity of $\pi^{\mathrm{l}}$ with respect to $Q^{\mathrm{l}}_{\pi^{\mathrm{h}},\pi^{\mathrm{l}}}$*

This directly follows from Assumption 3.

*2a) Lipschitz continuity of $G^2_{Q^{\mathrm{h}},P^{\mathrm{h}}_{\pi^{\mathrm{l}}}}$ with respect to $Q^{\mathrm{h}}$*

Starting from the definition of $G^2_{Q^{\mathrm{h}},P^{\mathrm{h}}_{\pi^{\mathrm{l}}}}(s^{\mathrm{h}},\omega^{\mathrm{h}})$:

$$
\begin{aligned}
&|G^2_{Q^{\mathrm{h},1},P^{\mathrm{h}}_{\pi^{\mathrm{l}}}}(s^{\mathrm{h}},\omega^{\mathrm{h}})-G^2_{Q^{\mathrm{h},2},P^{\mathrm{h}}_{\pi^{\mathrm{l}}}}(s^{\mathrm{h}},\omega^{\mathrm{h}})|\\
&=\left|(\gamma^{\mathrm{h}})^T\mathop{\mathbb{E}}_{s^{\mathrm{h},+}\sim P^{\mathrm{h}}_{\pi^{\mathrm{l}}}}\left[\max_{\omega^+}Q^{\mathrm{h},1}(s^{\mathrm{h},+},\omega^+)\right]-(\gamma^{\mathrm{h}})^T\mathop{\mathbb{E}}_{s^{\mathrm{h},+}\sim P^{\mathrm{h}}_{\pi^{\mathrm{l}}}}\left[\max_{\omega^+}Q^{\mathrm{h},2}(s^{\mathrm{h},+},\omega^+)\right]\right|\\
&=\left|(\gamma^{\mathrm{h}})^T\sum_{s^{\mathrm{h},+}\in\mathcal{S}^{\mathrm{h}}}P^{\mathrm{h}}_{\pi^{\mathrm{l}}}(s^{\mathrm{h},+}|s^{\mathrm{h}},\omega^{\mathrm{h}})\left[\max_{\omega^+}Q^{\mathrm{h},1}(s^{\mathrm{h},+},\omega^+)-\max_{\omega^+}Q^{\mathrm{h},2}(s^{\mathrm{h},+},\omega^+)\right]\right|\\
&\le(\gamma^{\mathrm{h}})^T\sum_{s^{\mathrm{h},+}\in\mathcal{S}^{\mathrm{h}}}P^{\mathrm{h}}_{\pi^{\mathrm{l}}}(s^{\mathrm{h},+}|s^{\mathrm{h}},\omega^{\mathrm{h}})\left|\max_{\omega^+}Q^{\mathrm{h},1}(s^{\mathrm{h},+},\omega^+)-\max_{\omega^+}Q^{\mathrm{h},2}(s^{\mathrm{h},+},\omega^+)\right|\\
&\le(\gamma^{\mathrm{h}})^T\sum_{s^{\mathrm{h},+}\in\mathcal{S}^{\mathrm{h}}}P^{\mathrm{h}}_{\pi^{\mathrm{l}}}(s^{\mathrm{h},+}|s^{\mathrm{h}},\omega^{\mathrm{h}})\left|\max_{\omega^+}\left(Q^{\mathrm{h},1}(s^{\mathrm{h},+},\omega^+)-Q^{\mathrm{h},2}(s^{\mathrm{h},+},\omega^+)\right)\right|\\
&\le(\gamma^{\mathrm{h}})^T\sum_{s^{\mathrm{h},+}\in\mathcal{S}^{\mathrm{h}}}P^{\mathrm{h}}_{\pi^{\mathrm{l}}}(s^{\mathrm{h},+}|s^{\mathrm{h}},\omega^{\mathrm{h}})\max_{\omega^+}\left|Q^{\mathrm{h},1}(s^{\mathrm{h},+},\omega^+)-Q^{\mathrm{h},2}(s^{\mathrm{h},+},\omega^+)\right|\\
&\le(\gamma^{\mathrm{h}})^T\sum_{s^{\mathrm{h},+}\in\mathcal{S}^{\mathrm{h}}}P^{\mathrm{h}}_{\pi^{\mathrm{l}}}(s^{\mathrm{h},+}|s^{\mathrm{h}},\omega^{\mathrm{h}})\max_{\bar{s}^{\mathrm{h}},\omega^+}\left|Q^{\mathrm{h},1}(\bar{s}^{\mathrm{h}},\omega^+)-Q^{\mathrm{h},2}(\bar{s}^{\mathrm{h}},\omega^+)\right|\\
&\le(\gamma^{\mathrm{h}})^T\underbrace{\max_{\bar{s}^{\mathrm{h}},\omega^+}\left|Q^{\mathrm{h},1}(\bar{s}^{\mathrm{h}},\omega^+)-Q^{\mathrm{h},2}(\bar{s}^{\mathrm{h}},\omega^+)\right|}_{=\|Q^{\mathrm{h},1}-Q^{\mathrm{h},2}\|_\infty}\underbrace{\sum_{s^{\mathrm{h},+}\in\mathcal{S}^{\mathrm{h}}}P^{\mathrm{h}}_{\pi^{\mathrm{l}}}(s^{\mathrm{h},+}|s^{\mathrm{h}},\omega^{\mathrm{h}})}_{=1}\\
&=(\gamma^{\mathrm{h}})^T\left\|Q^{\mathrm{h},1}-Q^{\mathrm{h},2}\right\|_\infty.
\end{aligned}
$$

The result easily follows.

*2b) Lipschitz continuity of $G^2_{Q^{\mathrm{h}},P^{\mathrm{h}}_{\pi^{\mathrm{l}}}}$ with respect to $P^{\mathrm{h}}_{\pi^{\mathrm{l}}}$*

The desired bound can be found as:

$$
\begin{aligned}
&|G^2_{Q^{\mathrm{h}},P^{\mathrm{h},1}}(s^{\mathrm{h}},\omega^{\mathrm{h}})-G^2_{Q^{\mathrm{h}},P^{\mathrm{h},2}}(s^{\mathrm{h}},\omega^{\mathrm{h}})|\\
&=(\gamma^{\mathrm{h}})^T\left|\sum_{s^{\mathrm{h},+}\in\mathcal{S}^{\mathrm{h}}}\left(P^{\mathrm{h},1}(s^{\mathrm{h},+}|s^{\mathrm{h}},\omega^{\mathrm{h}})-P^{\mathrm{h},2}(s^{\mathrm{h},+}|s^{\mathrm{h}},\omega^{\mathrm{h}})\right)\max_{\omega^+}Q^{\mathrm{h}}(s^{\mathrm{h},+},\omega^+)\right|\\
&\le(\gamma^{\mathrm{h}})^T\sum_{s^{\mathrm{h},+}\in\mathcal{S}^{\mathrm{h}}}\left|P^{\mathrm{h},1}(s^{\mathrm{h},+}|s^{\mathrm{h}},\omega^{\mathrm{h}})-P^{\mathrm{h},2}(s^{\mathrm{h},+}|s^{\mathrm{h}},\omega^{\mathrm{h}})\right|\left|\max_{\omega^+}Q^{\mathrm{h}}(s^{\mathrm{h},+},\omega^+)\right|\\
&\le(\gamma^{\mathrm{h}})^T\left|\max_{\bar{s}^{\mathrm{h}},\omega^+}Q^{\mathrm{h}}(\bar{s}^{\mathrm{h}},\omega^+)\right|\sum_{s^{\mathrm{h},+}\in\mathcal{S}^{\mathrm{h}}}\left|P^{\mathrm{h},1}(s^{\mathrm{h},+}|s^{\mathrm{h}},\omega^{\mathrm{h}})-P^{\mathrm{h},2}(s^{\mathrm{h},+}|s^{\mathrm{h}},\omega^{\mathrm{h}})\right|\\
&\le(\gamma^{\mathrm{h}})^T\left|\max_{\bar{s}^{\mathrm{h}},\omega^+}Q^{\mathrm{h}}(\bar{s}^{\mathrm{h}},\omega^+)\right|\left|\mathcal{S}^{\mathrm{h}}\right|\max_{\bar{s}^{\mathrm{h}}}\left|P^{\mathrm{h},1}(\bar{s}^{\mathrm{h}}|s^{\mathrm{h}},\omega^{\mathrm{h}})-P^{\mathrm{h},2}(\bar{s}^{\mathrm{h}}|s^{\mathrm{h}},\omega^{\mathrm{h}})\right|\\
&\le(\gamma^{\mathrm{h}})^T\left|\max_{\bar{s}^{\mathrm{h}},\omega^+}Q^{\mathrm{h}}(\bar{s}^{\mathrm{h}},\omega^+)\right|\mathcal{S}^{\mathrm{h}}\left|\,\left\|P^{\mathrm{h},1}-P^{\mathrm{h},2}\right\|_\infty\right..
\end{aligned}
$$

Lipschitz continuity easily follows.

*2c) Lipschitz continuity of $P^{\mathrm{h}}_{\pi^{\mathrm{l}}}$ with respect to $\pi^{\mathrm{l}}$*

Starting from the high-level dynamics in Eq. (2) with $s^{\mathrm{h}}=s_0$, $s^{\mathrm{h},+}=s_T$, we write:

$$
\begin{aligned}
&|P^{\mathrm{h}}_{\pi^{\mathrm{l},1}}(s^{\mathrm{h},+}|s^{\mathrm{h}},\omega^{\mathrm{h}})-P^{\mathrm{h}}_{\pi^{\mathrm{l},2}}(s^{\mathrm{h},+}|s^{\mathrm{h}},\omega^{\mathrm{h}})|\\
&=\Bigg|\sum_{\substack{s_1,\dots,s_{T-1}\in\mathcal{S}\\ a_0,\dots,a_{T-1}\in\mathcal{A}}}\prod_{i=1}^{T}P(s_i|s_{i-1},a_{i-1})\pi^{\mathrm{l},1}\left(a_{i-1}|s_{i-1},\omega^{\mathrm{h}},i-1\right)+\\
&\quad-\sum_{\substack{s_1,\dots,s_{T-1}\in\mathcal{S}\\ a_0,\dots,a_{T-1}\in\mathcal{A}}}\prod_{i=1}^{T}P(s_i|s_{i-1},a_{i-1})\pi^{\mathrm{l},2}\left(a_{i-1}|s_{i-1},\omega^{\mathrm{h}},i-1\right)\Bigg|\\
&=\Bigg|\sum_{\substack{s_1,\dots,s_{T-1}\in\mathcal{S}\\ a_0,\dots,a_{T-1}\in\mathcal{A}}}\prod_{i=1}^{T}P(s_i|s_{i-1},a_{i-1})\cdot\\
&\quad\cdot\left[\prod_{i=1}^{T}\pi^{\mathrm{l},1}\left(a_{i-1}|s_{i-1},\omega^{\mathrm{h}},i-1\right)-\prod_{i=1}^{T}\pi^{\mathrm{l},2}\left(a_{i-1}|s_{i-1},\omega^{\mathrm{h}},i-1\right)\right]\Bigg|\\
&\le\sum_{\substack{s_1,\dots,s_{T-1}\in\mathcal{S}\\ a_0,\dots,a_{T-1}\in\mathcal{A}}}\prod_{i=1}^{T}P(s_i|s_{i-1},a_{i-1})\cdot\\
&\quad\cdot\left|\prod_{i=1}^{T}\pi^{\mathrm{l},1}\left(a_{i-1}|s_{i-1},\omega^{\mathrm{h}},i-1\right)-\prod_{i=1}^{T}\pi^{\mathrm{l},2}\left(a_{i-1}|s_{i-1},\omega^{\mathrm{h}},i-1\right)\right|\\
&=\sum_{\substack{s_1,\dots,s_{T-1}\in\mathcal{S}\\ a_0,\dots,a_{T-1}\in\mathcal{A}}}\prod_{i=1}^{T}P(s_i|s_{i-1},a_{i-1})\Bigg|\sum_{j=1}^{T}\prod_{k=1}^{j-1}\pi^{\mathrm{l},1}\left(a_{k-1}|s_{k-1},\omega^{\mathrm{h}},k-1\right)\cdot\\
&\quad\cdot\left(\pi^{\mathrm{l},1}\left(a_{j-1}|s_{j-1},\omega^{\mathrm{h}},j-1\right)-\pi^{\mathrm{l},2}\left(a_{j-1}|s_{j-1},\omega^{\mathrm{h}},j-1\right)\right)\cdot\\
&\quad\cdot\prod_{k=j+1}^{T}\pi^{\mathrm{l},1}\left(a_{k-1}|s_{k-1},\omega^{\mathrm{h}},k-1\right)\Bigg|\\
&\le\sum_{\substack{s_1,\dots,s_{T-1}\in\mathcal{S}\\ a_0,\dots,a_{T-1}\in\mathcal{A}}}\prod_{i=1}^{T}P(s_i|s_{i-1},a_{i-1})\cdot\\
&\quad\cdot\left|\sum_{j=1}^{T}\left(\pi^{\mathrm{l},1}\left(a_{j-1}|s_{j-1},\omega^{\mathrm{h}},j-1\right)-\pi^{\mathrm{l},2}\left(a_{j-1}|s_{j-1},\omega^{\mathrm{h}},j-1\right)\right)\right|\\
&\le\sum_{\substack{s_1,\dots,s_{T-1}\in\mathcal{S}\\ a_0,\dots,a_{T-1}\in\mathcal{A}}}\prod_{i=1}^{T}P(s_i|s_{i-1},a_{i-1})\cdot\\
&\quad\cdot\sum_{j=1}^{T}\left|\pi^{\mathrm{l},1}\left(a_{j-1}|s_{j-1},\omega^{\mathrm{h}},j-1\right)-\pi^{\mathrm{l},2}\left(a_{j-1}|s_{j-1},\omega^{\mathrm{h}},j-1\right)\right|\\
&\le T\left\|\pi^{\mathrm{l},1}-\pi^{\mathrm{l},2}\right\|_\infty\sum_{\substack{s_1,\dots,s_{T-1}\in\mathcal{S}\\ a_0,\dots,a_{T-1}\in\mathcal{A}}}\prod_{i=1}^{T}P(s_i|s_{i-1},a_{i-1})\\
&=T\left\|\pi^{\mathrm{l},1}-\pi^{\mathrm{l},2}\right\|_\infty\sum_{a_0,\dots,a_{T-1}\in\mathcal{A}}\underbrace{\sum_{s_1,\dots,s_{T-1}\in\mathcal{S}}\prod_{i=1}^{T}P(s_i|s_{i-1},a_{i-1})}_{=1}\\
&=T|\mathcal{A}|^T\left\|\pi^{\mathrm{l},1}-\pi^{\mathrm{l},2}\right\|_\infty,
\end{aligned}
$$

and the result follows.

Thus combining the results above, we can conclude that $g$ is Lipschitz continuous with respect to both its arguments. □

**Lemma 5 (Properties of $M^{(1)}_{n+1}(x_n, y_n)$ and $M^{(2)}_{n+1}(x_n, y_n)$).** *$\{M^{(1)}_n\}$ and $\{M^{(2)}_n\}$ are martingale difference sequences with respect to the increasing σ-fields*

$$\mathcal{F}_n := \sigma\left(x_m, y_m, M^{(1)}_{n+1}, M^{(2)}_{n+1}, m \le n\right), \quad n \ge 0,$$

*satisfying*

$$\mathbb{E}\left[\|M^{(1)}_{n+1}\|_2^2 \,|\mathcal{F}_n\right] \le K^{(1)}\left(1 + \|x_n\|_2^2 + \|y_n\|_2^2\right),$$

$$\mathbb{E}\left[\|M^{(2)}_{n+1}\|_2^2 \,|\mathcal{F}_n\right] \le K^{(2)}\left(1 + \|x_n\|_2^2 + \|y_n\|_2^2\right),$$

*for $n \ge 0$ and for some constants $K^{(1)}, K^{(2)} > 0$.*

**Proof.** We will prove Lemma 5 for $M^{(2)}_{n+1}(x_n, y_n)$, and similar steps can be applied to $M^{(1)}_{n+1}(x_n, y_n)$.

Let us firstly show that $\{M^{(2)}_n\}$ is a martingale difference sequence w.r.t. $\mathcal{F}_n$, i.e., $\mathbb{E}\left[M^{(2)}_{n+1}(x_n, y_n)|\, \mathcal{F}_n\right] = 0$. Considering one component of $M^{(2)}_{n+1}(x_n, y_n)$, we write the conditional expectation as:

$$\begin{aligned}
&\mathbb{E}\left[M^{(2)}_{n+1,s^{\rm h},\omega^{\rm h}}(x_n, y_n)|\, \mathcal{F}_n\right] \\
&= \mathop{\mathbb{E}}_{\substack{s^{\rm h}_{t+1} \\ \tau_{[tT,tT+T-1]}}}\Bigg[(\gamma^{\rm h})^T \max_\omega Q^{{\rm h},n}\left(s^{\rm h}_{t+1}, \omega\right) - (\gamma^{\rm h})^T \mathop{\mathbb{E}}_{s^{\rm h,+}\sim P^{\rm h}_{\pi^{\rm l}}}\left[\max_{\omega^+} Q^{{\rm h},n}(s^{\rm h,+}, \omega^+)\right] \\
&\quad + \sum_{k=0}^{T-1} (\gamma^{\rm h})^k r(s_{tT+k}, a_{tT+k}) - r^{\rm h}_{\pi^{\rm l}}(s^{\rm h}, \omega^{\rm h})|\, \mathcal{F}_n\Bigg] \\
&= (\gamma^{\rm h})^T \mathop{\mathbb{E}}_{s^{\rm h}_{t+1}}\left[\max_\omega Q^{{\rm h},n}\left(s^{\rm h}_{t+1}, \omega\right) |\, \mathcal{F}_n\right] - (\gamma^{\rm h})^T \mathop{\mathbb{E}}_{s^{\rm h,+}}\left[\max_{\omega^+} Q^{{\rm h},n}(s^{\rm h,+}, \omega^+)|\, \mathcal{F}_n\right] \\
&\quad + \mathop{\mathbb{E}}_{\substack{a_{tT+k}\sim\pi^{\rm l}(\cdot|s^{\rm l}_{tT+k}) \\ s^{\rm l}_{tT+k+1}\sim P^{\rm l}_{\pi^{\rm h}}(\cdot|s^{\rm l}_{tT+k}, a_{tT+k})}}\left[\sum_{k=0}^{T-1} (\gamma^{\rm h})^k r(s_{tT+k}, a_{tT+k})|s_{tT} = s^{\rm h},\ \mathcal{F}_n\right] \\
&\quad - \mathop{\mathbb{E}}_{\substack{a_k\sim\pi^{\rm l}(\cdot|s^{\rm l}_k) \\ s^{\rm l}_{k+1}\sim P^{\rm l}_{\pi^{\rm h}}(\cdot|s^{\rm l}_k, a_k)}}\left[\sum_{k=0}^{T-1} (\gamma^{\rm h})^k r(s_k, a_k)|s_0 = s^{\rm h},\ \mathcal{F}_n\right] \\
&= 0,
\end{aligned} \tag{C.8}$$

where, considering two time steps $k_1 \le k_2$, we denote a trajectory of $\mathcal{M}$ by $\tau_{[k_1,k_2]} := (s_{k_1}, a_{k_1}, s_{k_1+1}, a_{k_1+1} \ldots, s_{k_2}, a_{k_2})$.

Let us now show that $\mathbb{E}\left[\|M^{(2)}_{n+1}\|_2^2 \,|\mathcal{F}_n\right] \le K^{(2)}(1 + \|x_n\|_2^2 + \|y_n\|_2^2)$ for $n \ge 0, K^{(2)} > 0$.

We can bound $\left|(\gamma^{\rm h})^T \max_\omega Q^{{\rm h},n}(s^{\rm h}_{t+1}, \omega) - (\gamma^{\rm h})^T \mathbb{E}_{s^{\rm h,+}\sim P^{\rm h}_{\pi^{\rm l}}}[\max_{\omega^+} Q^{{\rm h},n}(s^{\rm h,+}, \omega^+)]\right|$ as:

$$\begin{aligned}
&\left|(\gamma^{\rm h})^T \max_\omega Q^{{\rm h},n}\left(s^{\rm h}_{t+1}, \omega\right) - (\gamma^{\rm h})^T \mathop{\mathbb{E}}_{s^{\rm h,+}\sim P^{\rm h}_{\pi^{\rm l}}}\left[\max_{\omega^+} Q^{{\rm h},n}(s^{\rm h,+}, \omega^+)\right]\right| \\
&= \left|(\gamma^{\rm h})^T \max_\omega Q^{{\rm h},n}\left(s^{\rm h}_{t+1}, \omega\right) - (\gamma^{\rm h})^T \sum_{s^{\rm h,+}\in S^{\rm h}} P^{\rm h}_{\pi^{\rm l}}(s^{\rm h,+}|s^{\rm h}, \omega^{\rm h}) \max_{\omega^+} Q^{{\rm h},n}(s^{\rm h,+}, \omega^+)\right| \\
&\le (\gamma^{\rm h})^T \left|\max_\omega Q^{{\rm h},n}\left(s^{\rm h}_{t+1}, \omega\right)\right| + (\gamma^{\rm h})^T \left|\sum_{s^{\rm h,+}\in S^{\rm h}} P^{\rm h}_{\pi^{\rm l}}(s^{\rm h,+}|s^{\rm h}, \omega^{\rm h}) \max_{\omega^+} Q^{{\rm h},n}(s^{\rm h,+}, \omega^+)\right| \\
&\le (\gamma^{\rm h})^T \left|\max_{\bar s^{\rm h},\omega} Q^{{\rm h},n}(\bar s^{\rm h}, \omega)\right| + (\gamma^{\rm h})^T \left|\sum_{s^{\rm h,+}\in S^{\rm h}} P^{\rm h}_{\pi^{\rm l}}(s^{\rm h,+}|s^{\rm h}, \omega^{\rm h}) \max_{\bar s^{\rm h},\omega^+} Q^{{\rm h},n}(\bar s^{\rm h}, \omega^+)\right| \\
&= (\gamma^{\rm h})^T \left\|Q^{{\rm h},n}\right\|_\infty + (\gamma^{\rm h})^T \left\|Q^{{\rm h},n}\right\|_\infty \sum_{s^{\rm h,+}\in S^{\rm h}} P^{\rm h}_{\pi^{\rm l}}(s^{\rm h,+}|s^{\rm h}, \omega^{\rm h}) \\
&= 2(\gamma^{\rm h})^T \left\|Q^{{\rm h},n}\right\|_\infty,
\end{aligned}$$

and we can easily show that $\left|\sum_{k=0}^{T-1} (\gamma^{\rm h})^k r(s_{tT+k}, a_{tT+k}) - r^{\rm h}_{\pi^{\rm l}}(s^{\rm h}, \omega^{\rm h})\right| \le 2T\bar r$.

Thus we have:

$$\begin{aligned}
&\mathbb{E}\left[|\ M^{(2)}_{n+1,s^{\rm h},\omega^{\rm h}}(x_n, y_n)|^2 \ |\mathcal{F}_n\right] \\
&\le \mathbb{E}\left[4(\gamma^{\rm h})^{2T} \left\|Q^{{\rm h},n}\right\|_\infty^2 + 4T^2\bar r^2 + 8(\gamma^{\rm h})^T T\bar r \left\|Q^{{\rm h},n}\right\|_\infty \ |\mathcal{F}_n\right] \\
&= 4(\gamma^{\rm h})^{2T} \left\|Q^{{\rm h},n}\right\|_\infty^2 + 4T^2\bar r^2 + 8(\gamma^{\rm h})^T T\bar r \left\|Q^{{\rm h},n}\right\|_\infty \\
&\le 4(\gamma^{\rm h})^{2T} \left\|Q^{{\rm h},n}\right\|_\infty^2 + 4T^2\bar r^2 + 1 + \left(8(\gamma^{\rm h})^T T\bar r \left\|Q^{{\rm h},n}\right\|_\infty\right)^2 \\
&= 4\left(1 + 16T^2\bar r^2\right)(\gamma^{\rm h})^{2T} \left\|Q^{{\rm h},n}\right\|_\infty^2 + 4T^2\bar r^2 + 1 \\
&\le K_{s^{\rm h},\omega^{\rm h}}\left(1 + \|x_n\|_2^2 + \|y_n\|_2^2\right),
\end{aligned}$$

for an appropriate $K_{s^{\rm h},\omega^{\rm h}} > 0$. The result then follows. □

In the following, we provide the proofs of Lemma 1 and Lemma 2, which are needed to prove Theorem 1.

*Proof of Lemma 1.* In order to show that the ODE (14) has a globally asymptotically stable equilibrium $\lambda(y)$, it suffices to show that, with a $Q^{\rm h}$ held fixed as a constant parameter, $r^{\rm l}(s^{\rm l}, a) + \gamma^{\rm l}\, \mathbb{E}_{s^{\rm l,+}\sim P^{\rm l}_{\pi^{\rm h}}}[\max_{a^+} Q^{\rm l}(s^{\rm l,+}, a^+)]$ is a contraction with respect to $Q^{\rm l}$. In fact, Theorem 3.1 in [47] applies and guarantees the convergence to the fixed point of $r^{\rm l}(s^{\rm l}, a) + \gamma^{\rm l}\, \mathbb{E}_{s^{\rm l,+}\sim P^{\rm l}_{\pi^{\rm h}}}[\max_{a^+} Q^{\rm l}(s^{\rm l,+}, a^+)]$, which is guaranteed to exist and be unique by the Banach fixed-point theorem.

The contraction of $r^{\rm l}(s^{\rm l}, a) + \gamma^{\rm l}\, \mathbb{E}_{s^{\rm l,+}\sim P^{\rm l}_{\pi^{\rm h}}}[\max_{a^+} Q^{\rm l}(s^{\rm l,+}, a^+)]$ can be easily shown with the following steps:

$$\begin{aligned}
&\left|r^{\rm l}(s^{\rm l}, a) + \gamma^{\rm l} \mathop{\mathbb{E}}_{s^{\rm l,+}\sim P^{\rm l}_{\pi^{\rm h}}}\left[\max_{a^+} Q^{\rm l,1}(s^{\rm l,+}, a^+)\right] - r^{\rm l}(s^{\rm l}, a) - \gamma^{\rm l} \mathop{\mathbb{E}}_{s^{\rm l,+}\sim P^{\rm l}_{\pi^{\rm h}}}\left[\max_{a^+} Q^{\rm l,2}(s^{\rm l,+}, a^+)\right]\right| \\
&= \left|\gamma^{\rm l} \mathop{\mathbb{E}}_{s^{\rm l,+}\sim P^{\rm l}_{\pi^{\rm h}}}\left[\max_{a^+} Q^{\rm l,1}(s^{\rm l,+}, a^+)\right] - \gamma^{\rm l} \mathop{\mathbb{E}}_{s^{\rm l,+}\sim P^{\rm l}_{\pi^{\rm h}}}\left[\max_{a^+} Q^{\rm l,2}(s^{\rm l,+}, a^+)\right]\right| \\
&= \gamma^{\rm l}\left|\sum_{s^{\rm l,+}\in S^{\rm l}} P^{\rm l}_{\pi^{\rm h}}(s^{\rm l,+}|s^{\rm l}, a)\left[\max_{a^+} Q^{\rm l,1}(s^{\rm l,+}, a^+) - \max_{a^+} Q^{\rm l,2}(s^{\rm l,+}, a^+)\right]\right| \\
&\le \gamma^{\rm l} \sum_{s^{\rm l,+}\in S^{\rm l}} P^{\rm l}_{\pi^{\rm h}}(s^{\rm l,+}|s^{\rm l}, a)\left|\max_{a^+} Q^{\rm l,1}(s^{\rm l,+}, a^+) - \max_{a^+} Q^{\rm l,2}(s^{\rm l,+}, a^+)\right| \\
&\le \gamma^{\rm l} \sum_{s^{\rm l,+}\in S^{\rm l}} P^{\rm l}_{\pi^{\rm h}}(s^{\rm l,+}|s^{\rm l}, a)\left|\max_{a^+}\left[Q^{\rm l,1}(s^{\rm l,+}, a^+) - Q^{\rm l,2}(s^{\rm l,+}, a^+)\right]\right| \\
&\le \gamma^{\rm l} \sum_{s^{\rm l,+}\in S^{\rm l}} P^{\rm l}_{\pi^{\rm h}}(s^{\rm l,+}|s^{\rm l}, a)\left|\max_{\bar s^+,a^+}\left[Q^{\rm l,1}(\bar s^{\rm l,+}, a^+) - Q^{\rm l,2}(\bar s^{\rm l,+}, a^+)\right]\right| \\
&\le \gamma^{\rm l} \sum_{s^{\rm l,+}\in S^{\rm l}} P^{\rm l}_{\pi^{\rm h}}(s^{\rm l,+}|s^{\rm l}, a) \max_{\bar s^+,a^+}\left|Q^{\rm l,1}(\bar s^{\rm l,+}, a^+) - Q^{\rm l,2}(\bar s^{\rm l,+}, a^+)\right| \\
&\le \gamma^{\rm l}\left\|Q^{\rm l,1} - Q^{\rm l,2}\right\|_\infty \sum_{s^{\rm l,+}\in S^{\rm l}} P^{\rm l}_{\pi^{\rm h}}(s^{\rm l,+}|s^{\rm l}, a) \\
&= \gamma^{\rm l}\left\|Q^{\rm l,1} - Q^{\rm l,2}\right\|_\infty.
\end{aligned}$$

We need to prove the Lipschitz continuity of $\lambda(y)$, which requires us to show that the fixed point of $r^{\rm l}(s^{\rm l}, a) + \gamma^{\rm l}\, \mathbb{E}_{s^{\rm l,+}\sim P^{\rm l}_{\pi^{\rm h}}}[\max_{a^+} Q^{\rm l}(s^{\rm l,+}, a^+)]$ is Lipschitz in $Q^{\rm h}$. It suffices to show that the fixed point of $r^{\rm l}(s^{\rm l}, a) + \gamma^{\rm l}\, \mathbb{E}_{s^{\rm l,+}\sim P^{\rm l}_{\pi^{\rm h}}}[\max_{a^+} Q^{\rm l}(s^{\rm l,+}, a^+)]$ is Lipschitz in $P^{\rm l}_{\pi^{\rm h}}$, as Lipschitz continuity of $P^{\rm l}_{\pi^{\rm h}}$ with respect to $Q^{\rm h}$ can be easily derived from Eq. (5) and Assumption 3.

Firstly, as the reward is bounded by $\bar r$ and we are in an infinite horizon discounted setting, we can bound any low-level Q-function as:

$$|Q^{\rm l}_{\pi^{\rm h},\pi^{\rm l}}| \le \frac{\bar r}{1-\gamma} =: \bar Q^{\rm l}, \tag{C.9}$$

with $\bar Q^{\rm l} > 0$.

Let us now consider two fixed points of $r^{\rm l}(s^{\rm l}, a) + \gamma^{\rm l}\, \mathbb{E}_{s^{\rm l,+}\sim P^{\rm l}_{\pi^{\rm h}}}[\max_{a^+} Q^{\rm l}(s^{\rm l,+}, a^+)]$, namely $Q^{\rm l,1}$ and $Q^{\rm l,2}$, satisfying for all $s^{\rm l} \in S^{\rm l}, a \in \mathcal{A}$:

$$Q^{\rm l,1}(s^{\rm l}, a) = r^{\rm l}(s^{\rm l}, a) + \gamma^{\rm l} \mathop{\mathbb{E}}_{s^{\rm l,+}\sim P^{\rm l,1}}\left[\max_{a^+} Q^{\rm l,1}(s^{\rm l,+}, a^+)\right],$$

$$Q^{l,2}(s^l,a) = r^l(s^l,a) + \gamma^l \mathop{\mathbb{E}}_{s^{l,+}\sim P^{l,2}}\left[\max_{a^+} Q^{l,2}(s^{l,+},a^+)\right].$$

Now, for any $s^l \in \mathcal{S}^l, a \in \mathcal{A}$:

$$\begin{aligned}
&|Q^{l,1}(s^l,a) - Q^{l,2}(s^l,a)| \\
&\quad = \Big| r^l(s^l,a) + \gamma^l \mathop{\mathbb{E}}_{s^{l,+}\sim P^{l,1}}\left[\max_{a^+} Q^{l,1}(s^{l,+},a^+)\right] + \\
&\qquad - r^l(s^l,a) - \gamma^l \mathop{\mathbb{E}}_{s^{l,+}\sim P^{l,2}}\left[\max_{a^+} Q^{l,2}(s^{l,+},a^+)\right]\Big| \\
&\quad = \gamma^l \Big| \sum_{s^{l,+}} P^{l,1}(s^{l,+}|s^l,a)\max_{a^+} Q^{l,1}(s^{l,+},a^+) - \sum_{s^{l,+}} P^{l,2}(s^{l,+}|s^l,a)\max_{a^+} Q^{l,2}(s^{l,+},a^+)\Big| \\
&\quad \le \gamma^l \Big| \sum_{s^{l,+}} P^{l,1}(s^{l,+}|s^l,a)\max_{a^+} Q^{l,1}(s^{l,+},a^+) - \sum_{s^{l,+}} P^{l,2}(s^{l,+}|s^l,a)\max_{a^+} Q^{l,1}(s^{l,+},a^+)\Big| \\
&\quad + \gamma^l \Big| \sum_{s^{l,+}} P^{l,2}(s^{l,+}|s^l,a)\max_{a^+} Q^{l,1}(s^{l,+},a^+) - \sum_{s^{l,+}} P^{l,2}(s^{l,+}|s^l,a)\max_{a^+} Q^{l,2}(s^{l,+},a^+)\Big|.
\end{aligned} \tag{C.10}$$

We can now bound the first addend as:

$$\begin{aligned}
&\Big| \sum_{s^{l,+}} P^{l,1}(s^{l,+}|s^l,a)\max_{a^+} Q^{l,1}(s^{l,+},a^+) - \sum_{s^{l,+}} P^{l,2}(s^{l,+}|s^l,a)\max_{a^+} Q^{l,1}(s^{l,+},a^+)\Big| \\
&\quad = \Big| \sum_{s^{l,+}} \left(P^{l,1}(s^{l,+}|s^l,a) - P^{l,2}(s^{l,+}|s^l,a)\right)\max_{a^+} Q^{l,1}(s^{l,+},a^+)\Big| \\
&\quad \le \sum_{s^{l,+}} \left|P^{l,1}(s^{l,+}|s^l,a) - P^{l,2}(s^{l,+}|s^l,a)\right|\bar{Q}^l \\
&\quad \le \bar{Q}^l|\mathcal{S}^l| \left\|P^{l,1} - P^{l,2}\right\|_\infty.
\end{aligned} \tag{C.11}$$

Whereas the second addend can be bounded by:

$$\begin{aligned}
&\Big| \sum_{s^{l,+}} P^{l,2}(s^{l,+}|s^l,a)\max_{a^+} Q^{l,1}(s^{l,+},a^+) - \sum_{s^{l,+}} P^{l,2}(s^{l,+}|s^l,a)\max_{a^+} Q^{l,2}(s^{l,+},a^+)\Big| \\
&\quad = \Big| \sum_{s^{l,+}} P^{l,2}(s^{l,+}|s^l,a)\left(\max_{a^+} Q^{l,1}(s^{l,+},a^+) - \max_{a^+} Q^{l,2}(s^{l,+},a^+)\right)\Big| \\
&\quad \le \sum_{s^{l,+}} P^{l,2}(s^{l,+}|s^l,a)\Big|\max_{a^+} Q^{l,1}(s^{l,+},a^+) - \max_{a^+} Q^{l,2}(s^{l,+},a^+)\Big| \\
&\quad \le \left\|Q^{l,1} - Q^{l,2}\right\|_\infty \sum_{s^{l,+}} P^{l,2}(s^{l,+}|s^l,a) \\
&\quad = \left\|Q^{l,1} - Q^{l,2}\right\|_\infty.
\end{aligned} \tag{C.12}$$

Then, inserting Eqs. (C.11) and (C.12) in Eq. (C.10) we get for any $s^l \in \mathcal{S}^l, a \in \mathcal{A}$:

$$|Q^{l,1}(s^l,a) - Q^{l,2}(s^l,a)| \le \gamma^l\bar{Q}^l|\mathcal{S}^l| \left\|P^{l,1} - P^{l,2}\right\|_\infty + \gamma^l \left\|Q^{l,1} - Q^{l,2}\right\|_\infty. \tag{C.13}$$

From which we get:

$$\left\|Q^{l,1} - Q^{l,2}\right\|_\infty \le \frac{\gamma^l}{1-\gamma^l}\bar{Q}^l|\mathcal{S}^l| \left\|P^{l,1} - P^{l,2}\right\|_\infty. \tag{C.14}$$

Lipschitz continuity of $\lambda(y)$ follows. □

*Proof of Lemma 2.* With arguments similar to the ones used in the proof of Lemma 1, it suffices to show that $r^{h}_{\pi^l}(s^h,\omega^h) + (\gamma^h)^T \mathbb{E}_{s^{h,+}\sim P^h_{\pi^l}}[\max_{\omega^+} Q^h(s^{h,+},\omega^+)]$ is a contraction. In that case, the Banach fixed-point theorem guarantees the existence and uniqueness of a fixed point and Theorem 3.1 in [47] guarantees the convergence of the ODE (15) to the fixed point.

With similar steps to the ones in the proof of Lemma 1, it is easy to show that $r^{h}_{\pi^l}(s^h,\omega^h) + (\gamma^h)^T \mathbb{E}_{s^{h,+}\sim P^h_{\pi^l}}[\max_{\omega^+} Q^h(s^{h,+},\omega^+)]$ is a contraction. □

By assuming boundedness of the iterations, one could use the Lemmas 1, 2, 4, and 5 to verify the assumptions of Theorem 8.1 in [25] and guarantee almost sure convergence of $(x_n, y_n)$ to $(\lambda(y^*), y^*)$, which would translate, for a GLIE policy, into having $(Q^{l,n}, Q^{h,n}) \to (Q^{l,*}, Q^{h,*})$ a.s.

Instead of assuming boundedness of the iterations, we will use the results in [38] to guarantee it.

**Lemma 6 (Additional properties of $h$ for stability).** *The functions $h_c(x,y) := \frac{h(cx,cy)}{c}$, $c \ge 1$, converge compactly to some $h_\infty$. The limiting ODE $\frac{dx}{dt} = h_\infty(x(t),y)$ has a unique globally asymptotically stable equilibrium $\lambda_\infty(y)$, where $\lambda_\infty : \mathbb{R}^{|\mathcal{S}^h|\times|\Omega|} \to \mathbb{R}^{|\mathcal{S}^l|\times|\mathcal{A}|}$, is a Lipschitz map and satisfies $\lambda_\infty(0) = 0$.*

**Proof.** We want to show that, for any compact set $K \subset \mathbb{R}^{|\mathcal{S}^l|\times|\mathcal{A}|} \times \mathbb{R}^{|\mathcal{S}^h|\times|\Omega|}$:

$$\lim_{c\to\infty} \sup_{(Q^l,Q^h)\in K} |h_c(Q^l,Q^h) - h_\infty(Q^l,Q^h)| = 0. \tag{C.15}$$

We will work with a component of $h_c$, i.e., $h_{c,s^l,a} := \frac{h_{s^l,a}(cx,cy)}{c}$, and we will show that:

$$\lim_{c\to\infty} \sup_{(Q^l,Q^h)\in K} |h_{c,s^l,a}(Q^l,Q^h) - h_{\infty,s^l,a}(Q^l,Q^h)| = 0, \quad \forall s^l \in \mathcal{S}^l, a \in \mathcal{A}. \tag{C.16}$$

It is easy to see that if Eq. (C.16) holds then Eq. (C.15) holds as well.

It can easily be shown that $h_c$ and $h_\infty$ are:

$$\begin{aligned}
h_{c,s^l,a}(Q^l,Q^h) &= \frac{r^l(s^l,a)}{c} + \gamma^l \sum_{s^{l,+}\in\mathcal{S}^l} P^l_{\pi^h_c}(s^{l,+}|s^l,a)\max_{a^+} Q^l(s^{l,+},a^+) - Q^l(s^l,a) \\
h_{\infty,s^l,a}(Q^l,Q^h) &:= \gamma^l \sum_{s^{l,+}\in\mathcal{S}^l} P^l_{\pi^h_\infty}(s^{l,+}|s^l,a)\max_{a^+} Q^l(s^{l,+},a^+) - Q^l(s^l,a).
\end{aligned}$$

We want to find a bound on $|h_{c,s^l,a}(Q^l,Q^h) - h_{\infty,s^l,a}(Q^l,Q^h)|$:

$$\begin{aligned}
&|h_{c,s^l,a}(Q^l,Q^h) - h_{\infty,s^l,a}(Q^l,Q^h)| \\
&\quad = \Big| \frac{r^l(s^l,a)}{c} + \gamma^l \sum_{s^{l,+}\in\mathcal{S}^l} P^l_{\pi^h_c}(s^{l,+}|s^l,a)\max_{a^+} Q^l(s^{l,+},a^+) + \\
&\qquad - \gamma^l \sum_{s^{l,+}\in\mathcal{S}^l} P^l_{\pi^h_\infty}(s^{l,+}|s^l,a)\max_{a^+} Q^l(s^{l,+},a^+)\Big| \\
&\quad \le \Big| \frac{r^l(s^l,a)}{c} + \gamma^l \sum_{s^{l,+}\in\mathcal{S}^l} \left(P^l_{\pi^h_c}(s^{l,+}|s^l,a) - P^l_{\pi^h_\infty}(s^{l,+}|s^l,a)\right)\max_{a^+} Q^l(s^{l,+},a^+)\Big| \\
&\quad \le \Big| \frac{r^l(s^l,a)}{c}\Big| + \gamma^l \left\|Q^l\right\|_\infty |\mathcal{S}||\mathcal{S}^{de}| \sum_{\omega^+\in\Omega} \left|\pi^h_c(\omega^+|s) - \pi^h_\infty(\omega^+|s)\right|,
\end{aligned} \tag{C.17}$$

where $s$ is part of $s^l$.

Therefore, since $\lim_{c\to\infty} \frac{r^l(s^l,a)}{c} = 0$ and due to Assumption 7, $h_{c,s^l,a}$ converges uniformly on compacts.

Showing that $\frac{dx}{dt} = h_\infty(x(t),y)$ has a unique globally asymptotically stable equilibrium $\lambda_\infty(y)$ can be done similarly to the proof of Lemma 1. In this case, for a fixed $y = Q^h$, $\lambda_\infty(y)$ has as components the unique solution to:

$$0 = \gamma^l \sum_{s^{l,+}\in\mathcal{S}^l} P^l_{\pi^h}(s^{l,+}|s^l,a)\max_{a^+} Q^l(s^{l,+},a^+) - Q^l(s^l,a).$$

One can easily see that $\forall s^l \in \mathcal{S}^l, a \in \mathcal{A}$, $Q^l(s^l,a) = 0$ is the unique solution to the above equation for any fixed $y = Q^h$. This implies the required Lipschitz continuity condition and that $\lambda_\infty(0) = 0$. □

**Lemma 7 (Additional properties of $g$ for stability).** *The functions $g_c(x,y) := \frac{g(c\lambda_\infty(y),cy)}{c}$, $c \ge 1$, converge compactly to some $g_\infty$. The limiting*

*ODE* $\frac{dy}{dt} = g_\infty(\lambda_\infty(y(t)), y(t))$ *has the origin in* $\mathbb{R}^{|S^h|\times|\Omega|}$ *as its unique globally asymptotically stable equilibrium.*

**Proof.** The proof closely follows that of Lemma 6.

We want to show that, for any compact set $K \subset \mathbb{R}^{|S^l|\times|\mathcal{A}|} \times \mathbb{R}^{|S^h|\times|\Omega|}$:

$$\lim_{c\to\infty} \sup_{(Q^l,Q^h)\in K} |g_c(Q^l, Q^h) - g_\infty(Q^l, Q^h)| = 0. \tag{C.18}$$

We will work with a component of $g_c$, i.e., $g_{c,s^h,\omega} := \frac{g_{s^h,\omega}(cx,cy)}{c}$, and we will show that:

$$\lim_{c\to\infty} \sup_{(Q^l,Q^h)\in K} |g_{c,s^h,\omega}(Q^l, Q^h) - g_{\infty,s^h,\omega}(Q^l, Q^h)| = 0, \quad \forall s^h \in \mathcal{S}^h, \omega \in \Omega. \tag{C.19}$$

It is easy to see that if Eq. (C.19) holds then Eq. (C.18) holds as well.

It can easily be shown that $g_c$ and $g_\infty$ are:

$$\begin{aligned} g_{c,s^h,\omega}(Q^l, Q^h) = {} & \frac{r^h_{\pi^l_c}(s^l, a)}{c} + (\gamma^h)^T \sum_{s^{h,+}\in\mathcal{S}^h} P^h_{\pi^l_c}(s^{h,+}|s^h,\omega) \\ & \times \max_{\omega^+} Q^h(s^{h,+},\omega^+) - Q^h(s^h,\omega), \\ g_{\infty,s^h,\omega}(Q^l, Q^h) := {} & (\gamma^h)^T \sum_{s^{h,+}\in\mathcal{S}^h} P^h_{\pi^l_\infty}(s^{h,+}|s^h,\omega) \max_{\omega^+} Q^h(s^{h,+},\omega^+) - Q^h(s^h,\omega). \end{aligned}$$

We want to find a bound on $|g_{c,s^h,\omega}(Q^l, Q^h) - g_{\infty,s^h,\omega}(Q^l, Q^h)|$:

$$\begin{aligned} & |g_{c,s^h,\omega}(Q^l, Q^h) - g_{\infty,s^h,\omega}(Q^l, Q^h)| = \\ & = \Bigg| \frac{r^h_{\pi^l_c}(s^l, a)}{c} + (\gamma^h)^T \sum_{s^{h,+}\in\mathcal{S}^h} P^h_{\pi^l_c}(s^{h,+}|s^h,\omega) \max_{\omega^+} Q^h(s^{h,+},\omega^+) + \\ & - (\gamma^h)^T \sum_{s^{h,+}\in\mathcal{S}^h} P^h_{\pi^l_\infty}(s^{h,+}|s^h,\omega) \max_{\omega^+} Q^h(s^{h,+},\omega^+) \Bigg| \\ & \le \left| \frac{r^h_{\pi^l_c}(s^l, a)}{c} \right| + (\gamma^h)^T \left\| Q^h \right\|_\infty \sum_{s^{h,+}\in\mathcal{S}^h} \left| P^h_{\pi^l_c}(s^{h,+}|s^h,\omega) - P^h_{\pi^l_\infty}(s^{h,+}|s^h,\omega) \right| \\ & \le \left| \frac{1}{c} T \frac{1-(\gamma^h)^T}{1-\gamma^h} \bar{r} \right| + (\gamma^h)^T \left\| Q^h \right\|_\infty \sum_{s^{h,+}\in\mathcal{S}^h} \left| P^h_{\pi^l_c}(s^{h,+}|s^h,\omega) - P^h_{\pi^l_\infty}(s^{h,+}|s^h,\omega) \right|. \end{aligned} \tag{C.20}$$

With steps similar to the ones in the proof of Lemma 4 (Lipschitz continuity of $P^h_{\pi^l}$ with respect to $\pi^l$), it is possible to further bound $|g_{c,s^h,\omega}(Q^l, Q^h) - g_{\infty,s^h,\omega}(Q^l, Q^h)|$ as:

$$\begin{aligned} & |g_{c,s^h,\omega}(Q^l, Q^h) - g_{\infty,s^h,\omega}(Q^l, Q^h)| \\ & \le \left| \frac{1}{c} T \frac{1-(\gamma^h)^T}{1-\gamma^h} \bar{r} \right| + (\gamma^h)^T \left\| Q^h \right\|_\infty \cdot \\ & \cdot \sum_{s^{h,+}} \sum_{\substack{s_1,\dots,s_{T-1} \\ a_0,\dots,a_{T-1}}} \left| \sum_{j=1}^{T} \left( \pi^l_c\left(a_{j-1}|s_{j-1},\omega^h, j-1\right) - \pi^l_\infty\left(a_{j-1}|s_{j-1},\omega^h, j-1\right) \right) \right|, \end{aligned} \tag{C.21}$$

where $s_0 = s^h$. Therefore, since the first addend tends to zero as $c \to \infty$ and due to Assumption 7, $g_{c,s^h,\omega}$ converges uniformly on compacts.

Showing that $\frac{dy}{dt} = g_\infty(\lambda_\infty(y(t)), y(t))$ has the origin as its unique globally asymptotically stable equilibrium can be done similarly to the proof of Lemma 6. In this case, the unique solution to:

$$0 = (\gamma^h)^T \sum_{s^{h,+}\in\mathcal{S}^h} P^h_{\pi^l_\infty}(s^{h,+}|s^h,\omega) \max_{\omega^+} Q^h(s^{h,+},\omega^+) - Q^h(s^h,\omega),$$

is given by $Q^h(s^h,\omega) = 0, \forall s^h \in \mathcal{S}^h, \omega \in \Omega$. □

Lemmas 6 and 7, together with the Lemmas 1, 2, 4, and 5, verify the assumptions of Theorem 10 in [38] and guarantee boundedness of the iterations. Having guaranteed that the iterates remain bounded, Lemmas 1, 2, 4, and 5, guarantee that our updates verify the assumptions of Theorem 8.1 in [25] and guarantee almost sure convergence of $(x_n, y_n)$ to $(\lambda(y^*), y^*)$, which would translate, for a GLIE policy, into having $(Q^{l,n}, Q^{h,n}) \to (Q^{l,*}, Q^{h,*})$ a.s.

## Data availability

We have inserted a link in the article where it will be possible to find the code and data used.

## Author biography

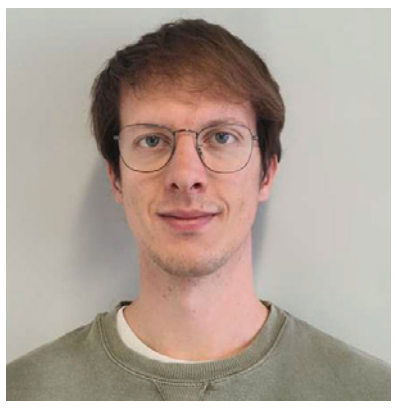

**Massimiliano Manenti** is a PhD student at the IST (Institute for Systems Theory and Automatic Control) at the University of Stuttgart. He received his B.Sc. and M.Sc. degrees in Automation and Control Engineering from the Polytechnic of Milan. His research interests include Hierarchical Reinforcement Learning, Optimal Control and Mobile Robotics.

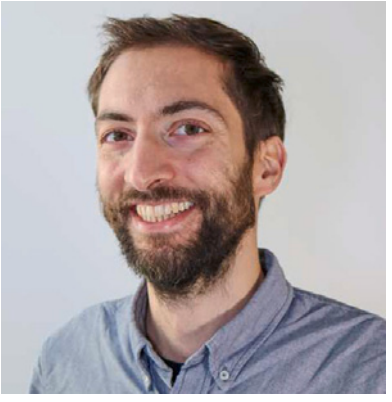

**Andrea Iannelli** is an Assistant Professor at the Institute for Systems Theory and Automatic Control at the University of Stuttgart. He completed his B.Sc. and M.Sc. degrees in Aerospace Engineering at the University of Pisa and received his PhD from the University of Bristol. He was also a post-doctoral researcher in the Automatic Control Laboratory at ETH Zurich. His research interests are centered around robust and adaptive control, uncertainty quantification, and sequential decision-making. He is an Associate Editor for the International Journal of Robust and Nonlinear Control and an IPC member for conferences in the areas of control, optimization, and learning.